\documentclass[10pt,twocolumn,letterpaper]{article}

\usepackage{cvpr}              
\usepackage{colortbl}      
\usepackage{multirow}
\usepackage{stfloats}
\definecolor{cvprblue}{rgb}{0.21,0.49,0.74}
\definecolor{myred}{HTML}{FD6864}   
\definecolor{myorange}{HTML}{FFCB2F}   
\definecolor{myyellow}{HTML}{FCFF2F}  
\usepackage[pagebackref,breaklinks,colorlinks,allcolors=cvprblue]{hyperref}

\def\paperID{} 
\def\confName{CVPR}
\def\confYear{2026}
\def\modelName{DualSplat}

\title{\modelName: Robust 3D Gaussian Splatting via Pseudo-Mask Bootstrapping from Reconstruction Failures}

\author{Xu Wang\textsuperscript{1} \quad 
Zhiru Wang\textsuperscript{1} \quad 
Shiyun Xie\textsuperscript{1} \quad 
Chengwei Pan\textsuperscript{1$\dagger$}\quad 
Yisong Chen\textsuperscript{2}\quad \\
\textsuperscript{1}{Beihang University}  \quad
\textsuperscript{2}{Peking University} 
}

\begin{document}

\twocolumn[{
\renewcommand\twocolumn[1][]{#1}
\maketitle
\begin{center}
    \vspace{-1em}
    \centering
    \includegraphics[width=\textwidth]{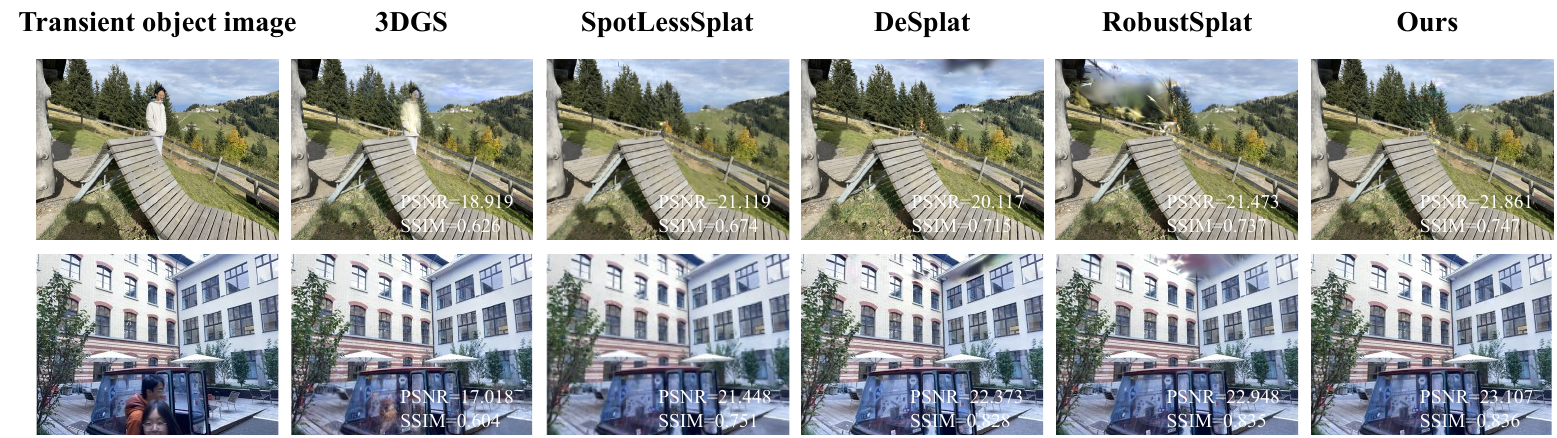}
    \captionof{figure}{\textbf{Transient objects in the training images introduce noticeable artifacts in the reconstruction results.} Compared with other methods, our approach achieves higher fidelity and more accurate suppression in transient regions.}
    \label{img1}
\end{center}%
}]

\maketitle

\renewcommand{\thefootnote}{\fnsymbol{footnote}}
\footnotetext[2]{Corresponding author.}
\renewcommand{\thefootnote}{\arabic{footnote}}

\begin{abstract}
While 3D Gaussian Splatting (3DGS) achieves real-time photorealistic rendering, its performance degrades significantly when training images contain transient objects that violate multi-view consistency. 
Existing methods face a circular dependency: accurate transient detection requires a well-reconstructed static scene, while clean reconstruction itself depends on reliable transient masks. 
We address this challenge with \modelName, a \textbf{Failure-to-Prior} framework that converts first-pass reconstruction failures into explicit priors for a second reconstruction stage. 
We observe that transients, which appear in only a subset of views, often manifest as incomplete fragments during conservative initial training.
We exploit these failures to construct object-level pseudo-masks by combining photometric residuals, feature mismatches, and SAM2 instance boundaries.
These pseudo-masks then guide a clean second-pass 3DGS optimization, while a lightweight MLP refines them online by gradually shifting from prior supervision to self-consistency. 
Experiments on RobustNeRF and NeRF On-the-go show that \modelName~ outperforms existing baselines, demonstrating particularly clear advantages in transient-heavy scenes and transient regions.
Project page: \href{https://lans1ot.github.io/DualSplat/}{https://lans1ot.github.io/DualSplat/}.

\end{abstract}

\section{Introduction}
\label{sec:intro}

3D Gaussian Splatting (3DGS) \cite{kerbl20233d} has emerged as a powerful paradigm for real-time photorealistic rendering. Its success, however, relies on a basic assumption: all training views should depict a static scene under mutually consistent observations. This assumption is often violated in real-world captures, where transient objects such as pedestrians, vehicles, and temporary occluders appear only in a subset of views. When this assumption is violated, 3DGS mistakenly absorbs these view-inconsistent observations into the scene representation by spawning spurious Gaussians, leading to ghosting artifacts and degraded reconstruction quality \cite{kulhanek2024wildgaussians}.

Existing approaches to transient-robust reconstruction mainly follow two directions. NeRF-based methods \cite{sabour2023robustnerf, ren2024nerf} suppress transients through uncertainty prediction or robust losses that down-weight inconsistent pixels, but they remain computationally expensive due to volumetric rendering. Recent 3DGS-based methods \cite{fu2025robustsplat, sabour2025spotlesssplats, lin2025hybridgs, wang2025desplat} improve efficiency by incorporating pretrained features, learned masks, or explicit static/transient decomposition. Despite their strong results, most of these methods still detect transients online while optimizing scene geometry, making transient suppression tightly coupled with reconstruction itself.

This coupling creates a fundamental circular dependency: accurate transient detection requires a well-reconstructed static scene to expose mismatches, yet clean reconstruction itself depends on reliable transient masks to prevent transient observations from being absorbed into geometry. When both processes are optimized jointly from poor initialization, errors reinforce one another: under-fitted static regions may be mistakenly suppressed as transients, while true transient content may instead be embedded into the reconstructed scene. Once such artifacts are baked into the geometry, the residual signal vanishes, making the failure nearly irreversible.

We address this problem by introducing a novel Failure-to-Prior paradigm. Our key observation is that transient objects, due to their sparse and inconsistent visibility across views, often appear as incomplete fragments as shown in \cref{fig:incomplete_fragments}. These failure patterns can be explicitly mined as cues for transient discovery. Specifically, we first perform a conservative 3DGS reconstruction to expose failure patterns, then convert these patterns into object-level pseudo-masks by consolidating photometric residuals, feature inconsistency, and SAM2 instance boundaries. These pseudo-masks serve as explicit priors for the second reconstruction stage, where a lightweight MLP further refines them online by gradually shifting from prior supervision to self-consistency as scene geometry stabilizes. In summary, our contributions are as follows:


\begin{figure}[t]
    \centering
    \includegraphics[width=\linewidth]{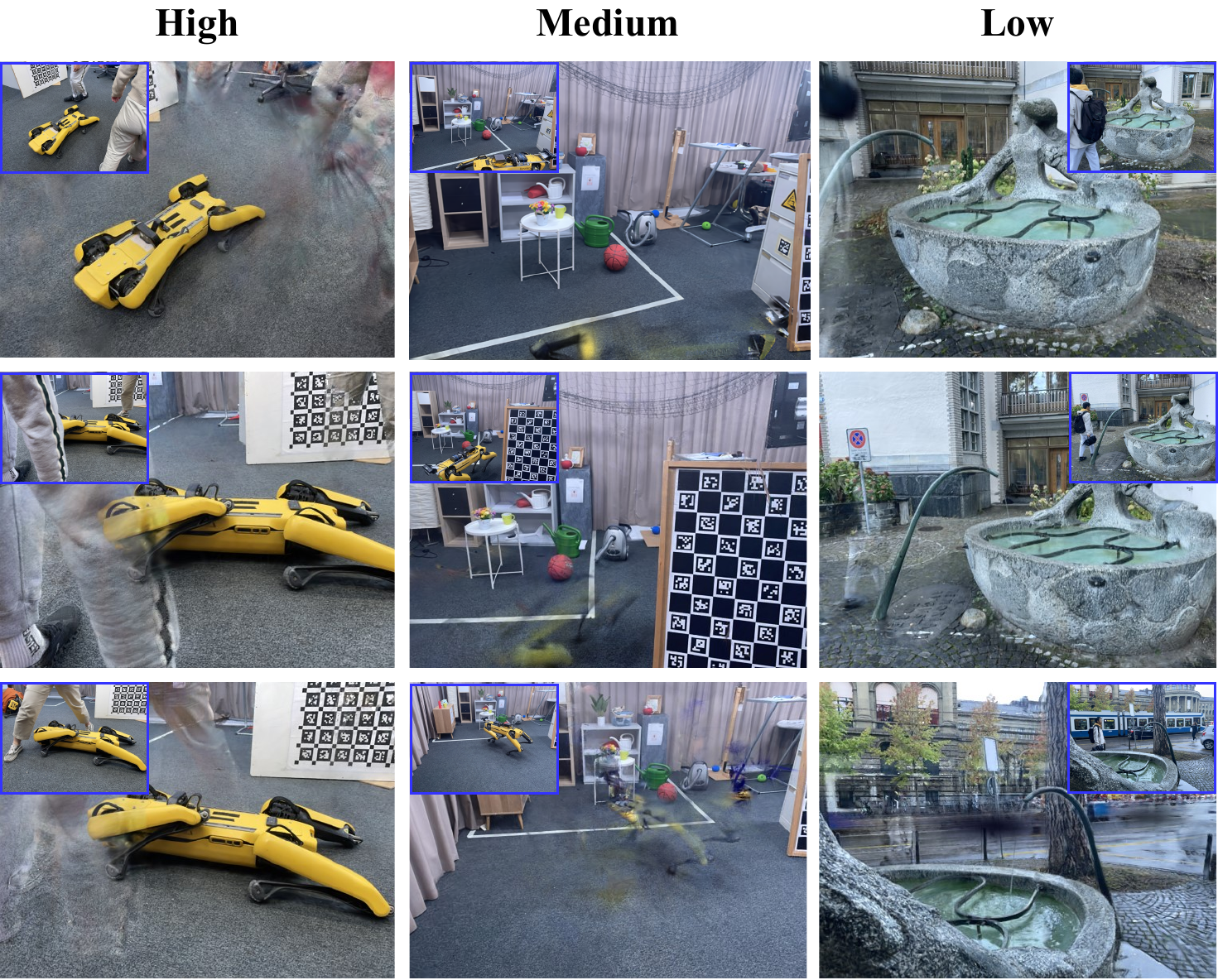}
    \caption{We select three scenes from the NeRF On-the-go dataset to showcase incomplete fragments, where “high”, “medium”, and “low” denote the proportion of dynamic objects in the scene.}
    \label{fig:incomplete_fragments}
    \vspace{-1em}
\end{figure}

\begin{enumerate}

    
    \item We propose a Failure-to-Prior paradigm for transient-robust 3DGS that breaks the circular dependency between transient detection and scene reconstruction by converting first-pass reconstruction failures into explicit priors.
    \item We introduce an unsupervised pseudo-mask construction pipeline that consolidates photometric residuals, view-consistent feature mismatches, and SAM2 instance boundaries into reliable object-level transient priors.
    \item We develop a prior-guided second-stage reconstruction framework with lightweight online mask refinement, where supervision gradually shifts from pseudo-mask priors to self-consistency as geometry stabilizes.
    \item We conduct comprehensive experiments on RobustNeRF and NeRF On-the-go, showing superior performance and robustness in transient-heavy scenes.
\end{enumerate}

\section{Related Work}

\subsection{Novel View Synthesis}

Early work represented scenes as implicit radiance fields rendered via volumetric integration, laying the foundation for neural view synthesis \cite{barron2021mip, barron2022mip, muller2022instant}. NeRF \cite{martin2021nerf} and its variants enabled high-quality novel view synthesis \cite{barron2021mip, barron2022mip, fridovich2022plenoxels, oechsle2021unisurf, xu2023grid}, and sparked a long line of improvements. More recently, 3DGS \cite{kerbl20233d} has attracted considerable attention as an effective alternative for novel view synthesis. Building on this foundation, 3DGS has emerged as a new backbone for tasks such as anti-aliasing \cite{yu2024mip, yan2024multi, liang2024analytic}, sparse-view reconstruction \cite{xiong2023sparsegs, xu2024mvpgs, zhang2024cor, zhu2024fsgs}, and generative modeling \cite{tang2023dreamgaussian, yi2024gaussiandreamer}.

\subsection{Robustness in Reconstruction}

For in-the-wild data, research on robust NeRFs \cite{martin2021nerf, chen2022hallucinated, yang2023cross} has focused mainly on modeling appearance variations and suppressing transient interference. NeRF-W \cite{martin2021nerf} represents the appearance per-image with latent codes and learns 2D uncertainty or outlier masks to reduce the weight of high-error pixels. RobustNeRF \cite{sabour2023robustnerf} takes a robust estimation view, deriving truncated masks from photometric residuals. NeRF On-the-go \cite{ren2024nerf} trains an uncertainty predictor on DINOv2 \cite{oquab2023dinov2} features and uses it to modulate per-pixel reconstruction losses.


Unlike NeRF, which uses a continuous implicit MLP-based representation, 3DGS employs a discrete explicit representation. WildGaussians \cite{kulhanek2024wildgaussians} augments 3DGS with robust DINO-based \cite{oquab2023dinov2} features and an integrated appearance-modeling module to handle occlusions and appearance changes. Wild-GS \cite{xu2024wild} adapts to unconstrained photo collections by aligning per-pixel appearance to local Gaussians via triplane sampling of reference images. Departing from purely color-residual strategies, SpotLessSplats \cite{sabour2025spotlesssplats} takes advantage of semantic features from a pretrained diffusion model \cite{rombach2022high, tang2023emergent} to extract feature maps offline prior to training, then uses clustering or a lightweight MLP to isolate structured outliers in feature space. RobustSplat \cite{fu2025robustsplat} introduces a delayed-densification schedule and detects transient objects by exploiting feature consistency. DeSplat \cite{wang2025desplat} decomposes a scene into a static part and a view-wise transient part using only photometric minimization, while HybridGS \cite{lin2025hybridgs} combines 3DGS with view-wise 2D image Gaussians to disentangle dynamic and static elements. DroneSplat \cite{tang2025dronesplat} leverages residual cues and a global instance tracker to achieve transient-free reconstruction of sparse views captured by drone.

\subsection{Pretrained Models}


Early convolutional networks, such as VGG \cite{simonyan2014very} and ResNet \cite{he2016deep}, provide stable mid-level semantic and texture representations. More recent pretrained models offer stronger and more transferable representations. DINO \cite{oquab2023dinov2, simeoni2025dinov3} produces self-supervised Vision Transformer features that capture object-level semantics and spatial correspondence, supporting tasks such as image retrieval, segmentation, and dense matching. SAM \cite{kirillov2023segment, ravi2024sam} is a promptable segmentation model trained on billion-mask datasets and is widely used for object and part segmentation. Depth Anything \cite{yang2024depth} is a large-scale monocular depth foundation model that delivers strong zero-shot relative and metric depth, providing robust depth priors across diverse scenes. FiT3D \cite{yue2024improving} extends such foundation features into 3D by enforcing view-consistent supervision, improving spatial coherence and semantic alignment in multi-view reconstruction. Other works \cite{qin2024langsplat, li2025langsplatv2, peng20243d, cen2025segment, zhou2024feature} build semantic or queryable 3D feature fields using DINO or SAM features to enable robust, feature-aware scene representations and downstream 3D editing.


\section{Method}

\subsection{Preliminaries}
3D Gaussian Splatting (3DGS) models a scene as a set of optimizable Gaussian primitives $\mathcal{G}=\{(\boldsymbol{\mu}_i,\boldsymbol{\Sigma}_i,o_i,\mathbf{c}_i)\}_{i=1}^N$, where each primitive has a 3D mean $\boldsymbol{\mu}\in\mathbb{R}^3$, covariance $\boldsymbol{\Sigma}\in\mathbb{R}^{3\times3}$ (parameterized by scaling $s_i$ and rotation $R_i$), opacity $o_i\in[0,1]$, and a view-dependent color $\mathbf{c_i}$ parameterized by spherical harmonics. Images are rendered by splatting the primitives into screen space, followed by alpha blending in front-to-back order.

Under a pinhole camera with view transform $\mathbf{W}$, each 3D Gaussian projects into a 2D Gaussian with mean $\boldsymbol{\mu}'\in\mathbb{R}^2$ and covariance $\boldsymbol{\Sigma}' = \mathbf{J}\,\mathbf{W}\,\boldsymbol{\Sigma}\,\mathbf{W}^\top \mathbf{J}^\top \in \mathbb{R}^{2\times 2},$ where $\mathbf{J}\in \mathbb{R}^{2\times 3}$ is the Jacobian of the affine approximation of the local projective mapping.

The depth-sorted 3D gaussians that overlap a pixel are composited in a front-to-back order using standard alpha blending. The Gaussian parameters are optimized by minimizing the photometric reconstruction loss between the rendered image and its reference image:
\begin{equation}
\mathcal{L} = (1-\lambda_\text{D-SSIM})\,\mathcal{L}_{1} + \lambda_\text{D-SSIM}\mathcal{L}_{\text{D-SSIM}},
\end{equation}
where $\lambda_\text{D-SSIM}$ balances pixel-wise $\mathcal{L}_1$ and perceptual D-SSIM terms. For handling transients, a common practice is to estimate a per-pixel binary mask $M$ and suppress transient regions during loss computation:
\begin{equation}
\mathcal{L}_{\text{masked}}=(1-\lambda_{\text{D-SSIM}})\,M\!\odot\!\mathcal{L}_{1}+\lambda_{\text{D-SSIM}}\,M\!\odot\!\mathcal{L}_{\text{D-SSIM}},
\label{eq:loss}
\end{equation}
where $\odot$ denotes element-wise multiplication.

\begin{figure*}[!h]
    \centering
    \includegraphics[width=0.9\textwidth]{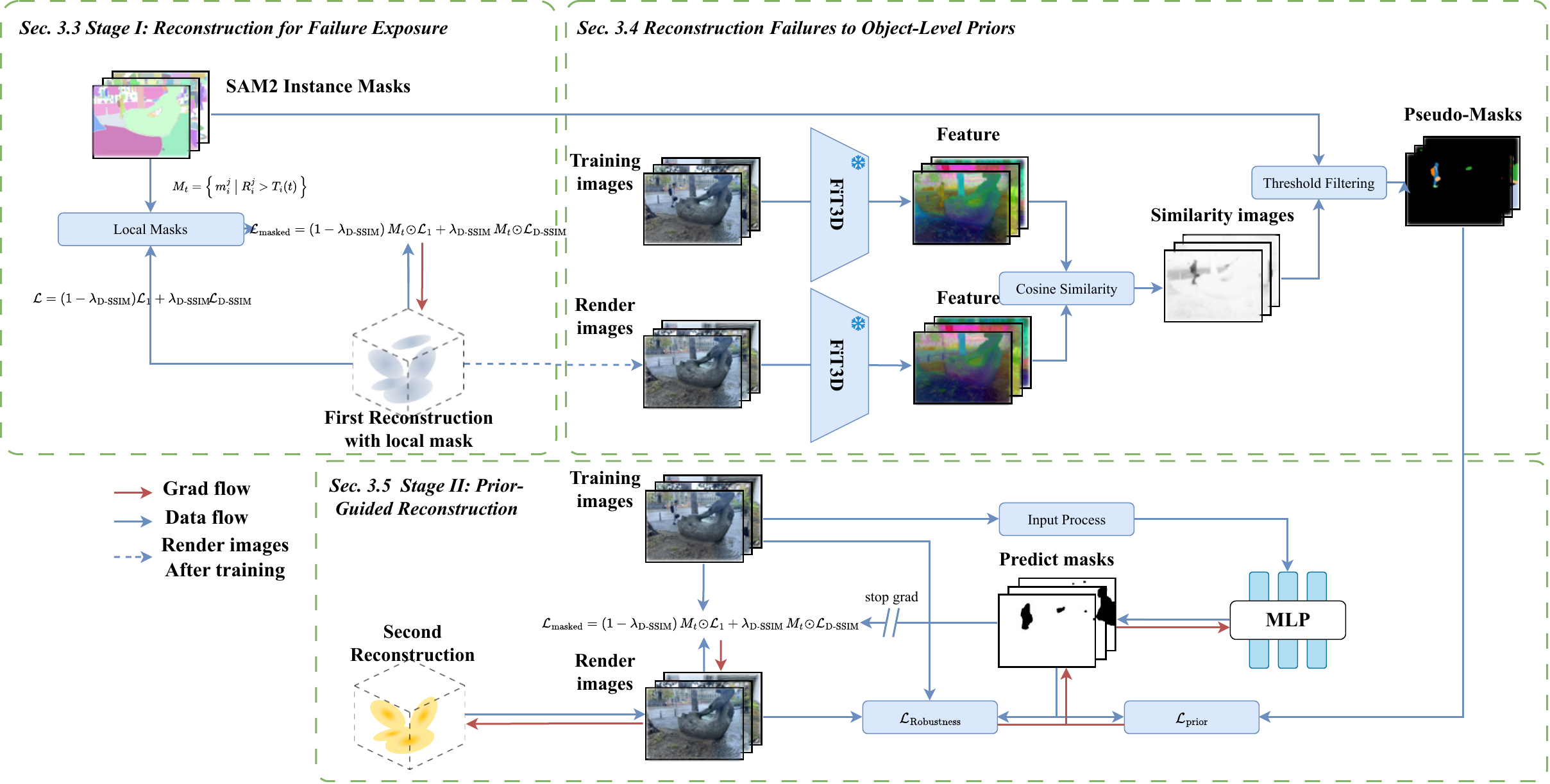}
    \caption{\modelName~performs two-stage 3D Gaussian Splatting to suppress transient distractions. The first stage reconstructs a coarse static scene. After the first training, Mask Filter produces confidence-weighted pseudo-masks. In the second stage, a lightweight MLP learns from the masks online using feature and depth cues, yielding final masks for reconstruction.
    }
    \label{method:overview}
\end{figure*}


\subsection{Overview}
Our method is built on a \textbf{Failure-to-Prior} principle: 
reconstruction failures caused by view-inconsistent transients are not merely artifacts to suppress, but signals that can be mined into priors. 
Instead of detecting transients online while scene geometry remains unstable, we first perform a conservative 3DGS reconstruction to expose failure patterns. We then convert these failures into object-level pseudo-masks to guide the second reconstruction.
This decouples transient discovery from clean scene optimization and avoids the signal-erasure failure mode of purely online heuristics. Compared in \cref{tab:comparision} with online suppression, our key difference is not the type of residual or feature cues, but \emph{when} and \emph{how} these cues are used. 

\begin{table}[h]
\centering
\vspace{-0.5em}
\caption{Comparison of paradigms and mechanisms.}
\vspace{-1em}
\label{tab:comparision}
\renewcommand{\arraystretch}{1.05}
\resizebox{\linewidth}{!}{
    \begin{tabular}{lcc}
    \toprule
    Item & Online suppression methods & Ours(\modelName) \\
    \midrule
    Paradigm & Online Heuristic (Internal) & Failure-to-Prior (External Guidance) \\
    Dependency & Circular (Entangled) & Decoupled (Sequential Stages) \\
    Priors & None (Internal Residuals only)	 & Instance (SAM2) + 3D Consistency (FiT3D)\\
    Failure Risk & Signal Loss from overfitting & Robust (Priors prevent overfitting)\\
    \bottomrule
    \end{tabular}
}
\vspace{-1em}
\end{table}

In our pipeline shown in \cref{method:overview}, the first reconstruction externalizes transient evidence into explicit priors before the final reconstruction begins.

\subsection{Stage I: Reconstruction for Failure Exposure}
\label{sec:stage1}

Rather than attempting a perfect separation of transient objects at the outset, Stage I performs an initial reconstruction to expose reconstruction failures. 

As illustrated in \cref{fig:incomplete_fragments}, transient objects present in training images often produce incomplete, blurred fragments in 3DGS reconstructions due to their sparse and inconsistent visibility across views. These fragments leak glimpses of the static background, creating photometric discrepancies that can be exploited to locate transients. However, while such artifacts are human-discernible, some of them can be challenging for deep neural networks to identify. To mitigate this identification challenge, we use a local masking strategy that requires minimal modifications to the standard 3DGS pipeline.

\textbf{Instance-level residual screening.}
We begin by training an initial 3DGS model and comparing each rendered image with its ground-truth training view. Inspired by RobustNeRF \cite{sabour2023robustnerf} and DroneSplat \cite{tang2025dronesplat}, we use photometric residuals as the first cue for transient discovery. Instead of thresholding pixels independently, we aggregate residuals at the object level using SAM2 instance masks
$S(I_i)=\{m_i^j\}_{j=1}^{N_i}$,
where $N_i$ is the number of instances in image $I_i$.
This instance-level formulation is more robust than pixel-wise masking, since transient corruption typically forms spatially coherent patterns within semantic objects.

For each instance mask $m_i^j$, we compute the spatially averaged residual
\begin{equation}
R_i^j=\frac{\sum_{p\in m_i^j} R(p)}{|m_i^j|},
\end{equation}
where $R(p)$ denotes the photometric residual at pixel $p$ and $|m_i^j|$ is the mask area.
For the current frame, we further compute the image-level residual statistics
\begin{equation}
\mu_i=\frac{1}{P}\sum_{p\in P} R(p), \qquad
\sigma_i^2=\frac{1}{P}\sum_{p\in P}(R(p)-\mu_i)^2,
\end{equation}
where $P$ is the set of all pixels in the image.

To account for the changing reliability of reconstruction over training, we adopt an adaptive threshold
\begin{equation}
T_i(t)=\mu_i+\left(1+\lambda_{\text{local}}\frac{T_{\max}-t}{T_{\max}}\right)\sigma_i,
\end{equation}
where $t$ denotes the current iteration and $T_{\max}$ is the total number of iterations.
This threshold is intentionally conservative in early training: when geometry is still unstable, a higher threshold prevents under-fitted static regions from being over-masked. As reconstruction improves, the threshold gradually tightens, allowing more transient candidates to be exposed. We mark an instance as transient when
\begin{equation}
M_t=\{m_i^j \mid R_i^j > T_i(t)\}.
\end{equation}
The resulting mask $M_t$ is then applied in the masked reconstruction loss of \cref{eq:loss}.

\subsection{Reconstruction Failures to Object-Level Priors}
\label{sec:prior}

The primary objective of this step is to translate these first-pass failures into reliable object-level priors for the second reconstruction stage, rather than directly outputting final transient masks.
Following our principle, we externalize transient evidence before the final reconstruction begins, rather than relying on the same optimization loop to both discover and suppress transients.

\begin{figure}[t]
    \centering
    \includegraphics[width=\linewidth]{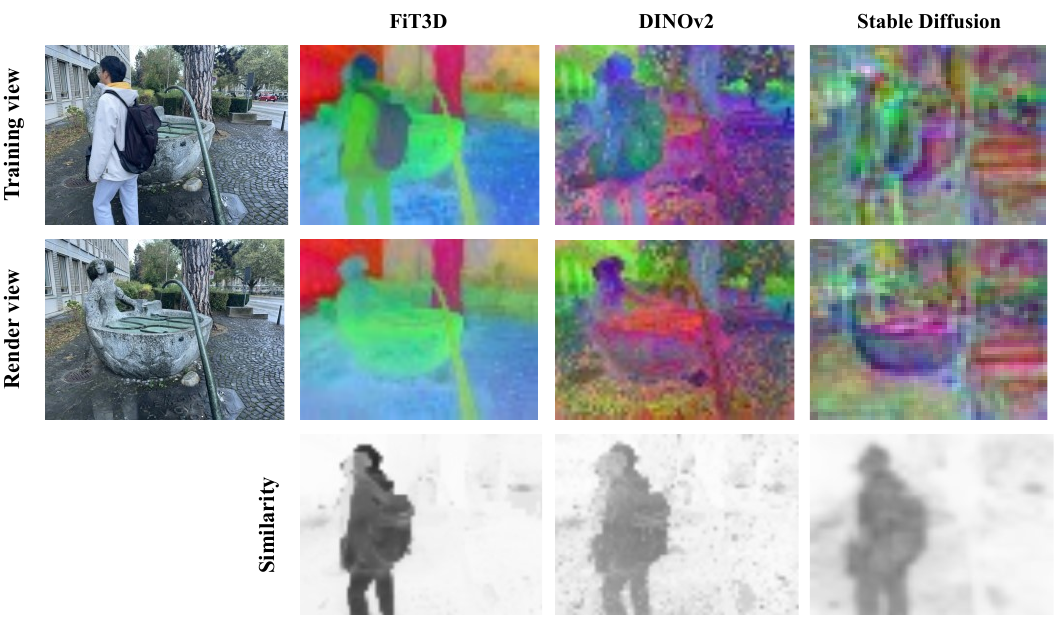}
    \caption{\textbf{Visualization results using different pretrained models as feature extractors.} FiT3D, when used as the feature extractor, produces the most distinct feature visualization and accurately identifies transient objects.}
    \label{fig:feature_compare}
    \vspace{-1em}
\end{figure}

We extract dense features from the ground-truth and first-pass rendered images using FiT3D\cite{yue2024improving}, which provides stronger view-consistent cues than generic image features as shown in \cref{fig:feature_compare}. 
Given the feature maps $F_{\text{gt}}$ and $F_{\text{render}}$, we compute a pixel-wise cosine-similarity map
\begin{equation}
S = \cos(F_{\text{gt}}, F_{\text{render}}).
\end{equation}
Low similarity indicates view-inconsistent regions that are likely caused by transient objects.

To obtain cleaner object-level priors, we aggregate both feature and photometric cues within SAM2 \cite{ravi2024sam} instance proposals. 
Let $\hat{S}\in[0,1]^{H\times W}$ be the min-max normalized similarity map. 
For an instance mask $m$ with pixel set $\Omega_m$, we compute
\begin{equation}
\mu_m=\frac{1}{|\Omega_m|}\sum_{p\in\Omega_m}\hat{S}(p),\qquad
\bar{\ell}_m=\frac{1}{|\Omega_m|}\sum_{p\in\Omega_m}L_1(p).
\end{equation}
where $L_1$ denotes per-pixel L1 residual. We retain $m$ as a transient prior only when it simultaneously exhibits low feature consistency and high photometric error, i.e.,
\begin{equation}
\mu_m \le \tau_{\text{sim}}, \qquad \bar{\ell}_m \ge \tau_{L1}.
\end{equation}
After a slight morphological dilation, the retained instances form the pseudo-masks $M_{\text{pseudo}}$.

These pseudo-masks are intentionally \emph{high-recall} rather than perfectly precise: over-masking some static regions is preferable to letting transients leak into the reconstructed geometry. 
The resulting priors provide external guidance that purely online residual-based methods lack, and are later learned and  refined in Stage~II once the scene geometry becomes more stable.

\begin{table*}[t] \centering
    \begin{center}\captionof{table}{\textbf{Comparison on NeRF On-the-go Dataset.} For better visualization, the \colorbox{myred}{1st}, \colorbox{myorange}{2nd} and \colorbox{myyellow}{3rd} best results are highlighted. Our method not only effectively eliminates dynamic distractors but also reconstructs fine details in the scene.}
\resizebox{\textwidth}{!}{
    \begin{tabular}{l|ccc|ccc|ccc|ccc|ccc|ccc|ccc}
    \toprule
    \multirow{3}{*}{Method} 
    & \multicolumn{6}{c|}{Low Occlusion}  
    & \multicolumn{6}{c|}{Medium Occlusion} 
    & \multicolumn{6}{c|}{High Occlusion} 
    & \multicolumn{3}{c}{\multirow{2}{*}{Mean}}  
    \\
    \multicolumn{1}{l|}{}  
    & \multicolumn{3}{c}{Mountain}  
    & \multicolumn{3}{c|}{Fountain} 
    & \multicolumn{3}{c}{Corner}  
    & \multicolumn{3}{c|}{Patio} 
    & \multicolumn{3}{c}{Spot} 
    & \multicolumn{3}{c|}{Patio-High}  
    \\
     & PSNR & SSIM & LPIPS 
     & PSNR & SSIM & LPIPS 
     & PSNR & SSIM & LPIPS 
     & PSNR & SSIM & LPIPS 
     & PSNR & SSIM & LPIPS 
     & PSNR & SSIM & LPIPS 
     & PSNR & SSIM & LPIPS 
     \\
    \midrule
    3DGS~\cite{kerbl20233d}
    & 18.92 & 0.626 & 0.179
    & 19.95 & 0.652 & 0.154
    & 23.11 & 0.840 & 0.103
    & 16.23 & 0.697 & 0.182
    & 19.03 & 0.764 & 0.263
    & 17.02 & 0.604 & 0.295
    & 19.04 & 0.697 & 0.196\\
    
    WildGaussians~\cite{kulhanek2024wildgaussians}
    & 20.73 & 0.707 & 0.190
    & 20.93 & 0.680 & 0.175
    & 25.61 & 0.878 & 0.072
    & \cellcolor{myorange}21.59 & 0.818 & 0.100
    & 24.59 & 0.877 & 0.073
    & 22.63 & 0.808 & 0.127
    & 22.68 & 0.795 & 0.123\\

    DeSplat~\cite{wang2025desplat}
    & 20.12 & 0.715 & 0.172
    & 20.31 & 0.675 & 0.166
    & 26.18 & 0.879 & 0.083
    & 19.94 & 0.800 & 0.125
    & 24.96 & 0.878 & 0.116
    & 22.37 & 0.828 & 0.134
    & 22.32 & 0.796 & 0.133\\
    
    SpotLessSplats~\cite{sabour2025spotlesssplats}
    & 21.12 & 0.674 & 0.204
    & 20.69 & 0.656 & 0.179
    & 25.71 & 0.860 & 0.083
    & \cellcolor{myyellow}21.54 & 0.804 & 0.095
    & 22.80 & 0.803 & 0.134
    & 21.48 & 0.760 & 0.167
    & 22.22 & 0.759 & 0.144\\

    DeGauss~\cite{wang2025degauss}
    & \cellcolor{myred}22.12 & \cellcolor{myred}0.754 & \cellcolor{myyellow}0.135
    & 20.74 & 0.677 & 0.149
    & \cellcolor{myyellow}26.32 & \cellcolor{myyellow}0.884 & \cellcolor{myred}0.049
    & 21.29 & 0.816 & 0.084 
    & \cellcolor{myred}26.00 & 0.875 & \cellcolor{myyellow}0.051
    & \cellcolor{myorange}23.09 & 0.820 & \cellcolor{myyellow}0.095
    & \cellcolor{myorange}23.26 & 0.804 & 0.094 \\

    RobustSplat~\cite{fu2025robustsplat}
    & \cellcolor{myyellow}21.47 & \cellcolor{myyellow}0.737 & 0.139
    & \cellcolor{myorange}21.19 & \cellcolor{myred}0.711 & \cellcolor{myorange}0.127
    & \cellcolor{myred}26.60 & \cellcolor{myred}0.898 & \cellcolor{myorange}0.049
    & 21.52 & \cellcolor{myorange}0.827 & \cellcolor{myorange}0.080
    & \cellcolor{myyellow}25.23 & \cellcolor{myorange}0.904 & \cellcolor{myorange}0.048
    & \cellcolor{myyellow}22.95 & \cellcolor{myorange}0.835 & \cellcolor{myred}0.094
    & \cellcolor{myyellow}23.16 & \cellcolor{myorange}0.819 & \cellcolor{myorange}0.089\\
    
    3DGS with pseudo-masks
    & 21.01 & 0.728 & \cellcolor{myorange}0.128
    & \cellcolor{myyellow}21.10 & \cellcolor{myyellow}0.697 & \cellcolor{myred}0.121
    & 24.96 & \cellcolor{myyellow}0.882 & 0.056
    & 21.33 & 0.822 & \cellcolor{myred}0.078
    & 24.52 & \cellcolor{myyellow}0.901 & 0.053
    & 22.71 & \cellcolor{myyellow}0.829 & 0.098
    & 22.60 & \cellcolor{myyellow}0.810 & \cellcolor{myyellow}0.089\\
 
    Ours 
    & \cellcolor{myorange}21.86 & \cellcolor{myorange}0.747 & \cellcolor{myred}0.127
    & \cellcolor{myred}21.28 & \cellcolor{myorange}0.706 & \cellcolor{myyellow}0.130
    & \cellcolor{myorange}26.53 & \cellcolor{myorange}0.897 & \cellcolor{myyellow}0.053
    & \cellcolor{myred}21.64 & \cellcolor{myred}0.828 & \cellcolor{myyellow}0.080
    & \cellcolor{myorange}25.96 & \cellcolor{myred}0.908 & \cellcolor{myred}0.046
    & \cellcolor{myred}23.26 & \cellcolor{myred}0.836 & \cellcolor{myorange}0.095
    & \cellcolor{myred}23.42 & \cellcolor{myred}0.820 & \cellcolor{myred}0.088\\
    \bottomrule
    \end{tabular}}
        \label{table:on-the-go}
    \end{center}
    \vspace{-15pt}
    \makebox[0.16\textwidth]{\footnotesize GT}
\makebox[0.16\textwidth]{\footnotesize DeSplat}
\makebox[0.16\textwidth]{\footnotesize WildGaussians}
\makebox[0.16\textwidth]{\footnotesize SpotLessSplats}
\makebox[0.16\textwidth]{\footnotesize RobustSplat}
\makebox[0.16\textwidth]{\footnotesize Ours}
\\
\includegraphics[width=0.16\textwidth]{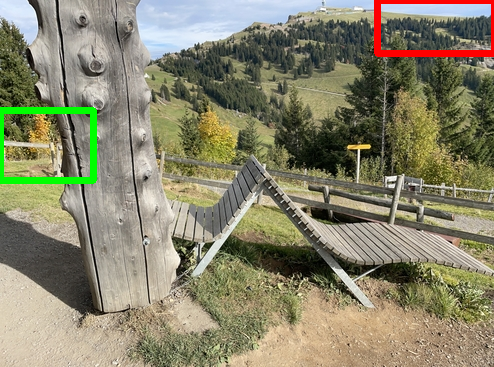}
\includegraphics[width=0.16\textwidth]{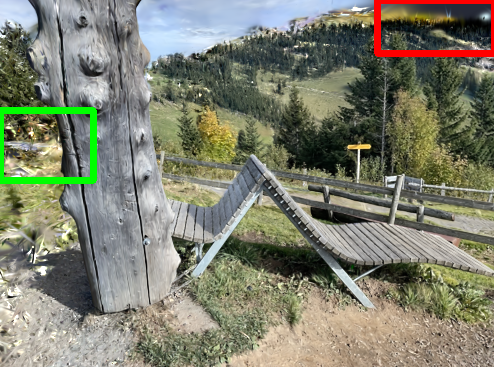}
\includegraphics[width=0.16\textwidth]{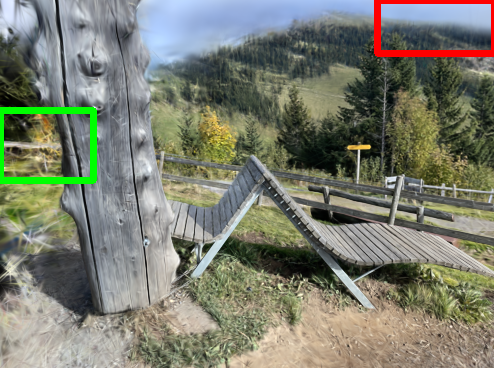}
\includegraphics[width=0.16\textwidth]{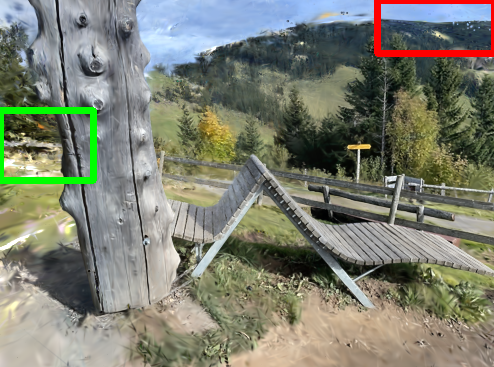}
\includegraphics[width=0.16\textwidth]{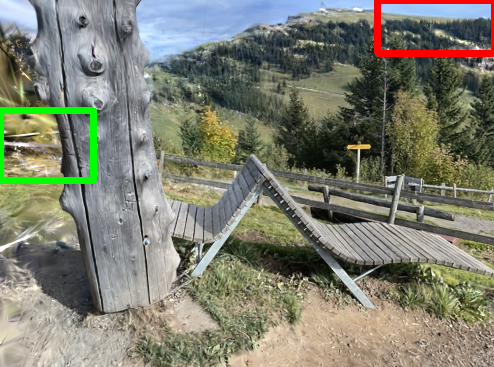}
\includegraphics[width=0.16\textwidth]{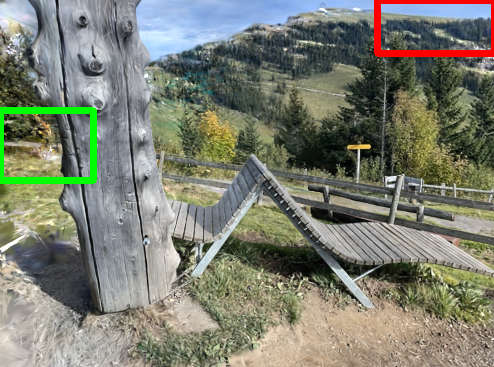}
\\
\includegraphics[width=0.0775\textwidth]{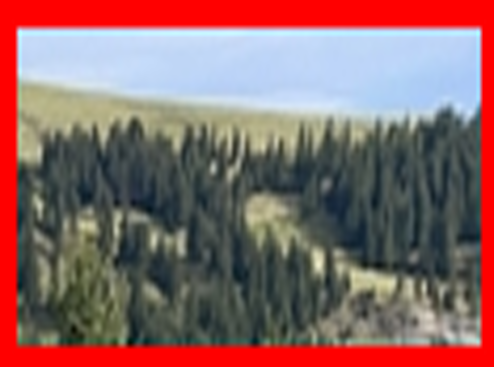}
\includegraphics[width=0.0775\textwidth]{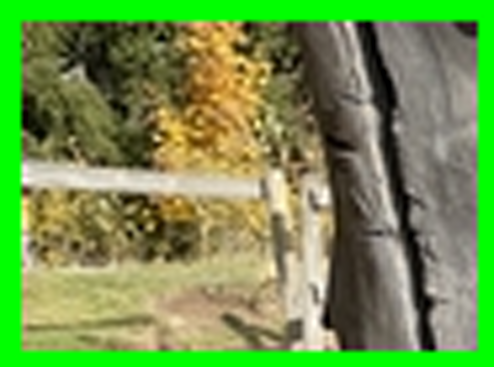}
\includegraphics[width=0.0775\textwidth]{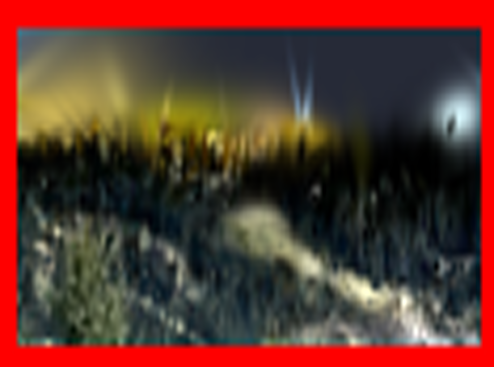}
\includegraphics[width=0.0775\textwidth]{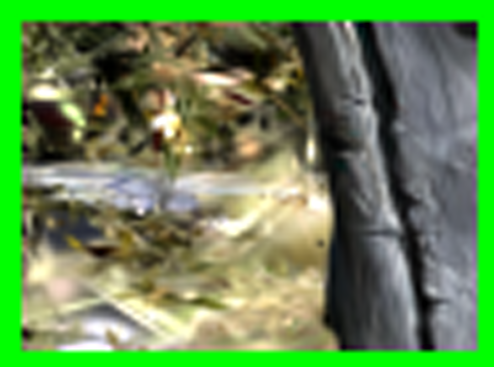}
\includegraphics[width=0.0775\textwidth]{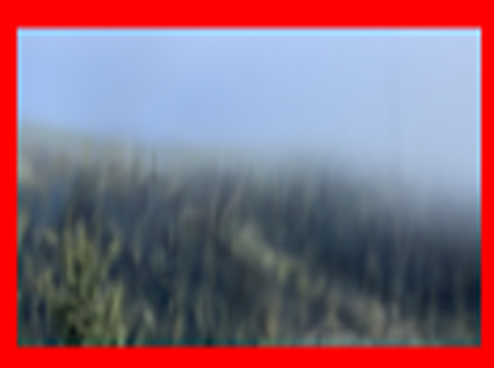}
\includegraphics[width=0.0775\textwidth]{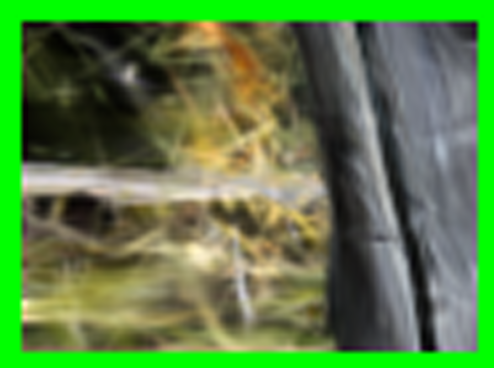}
\includegraphics[width=0.0775\textwidth]{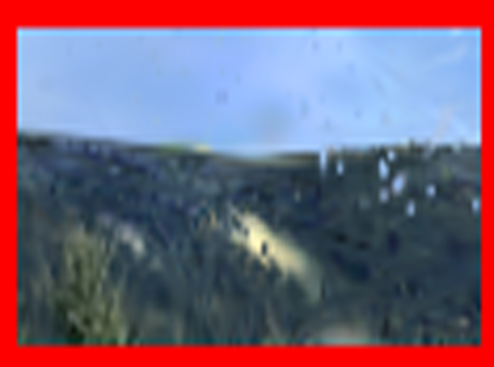}
\includegraphics[width=0.0775\textwidth]{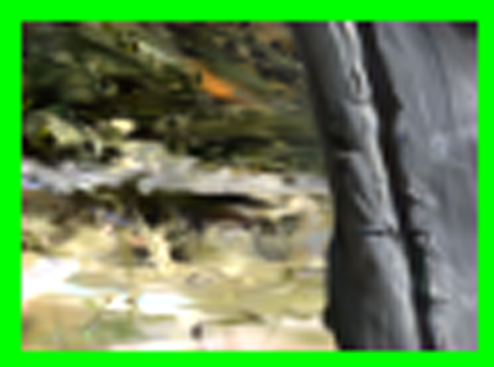}
\includegraphics[width=0.0775\textwidth]{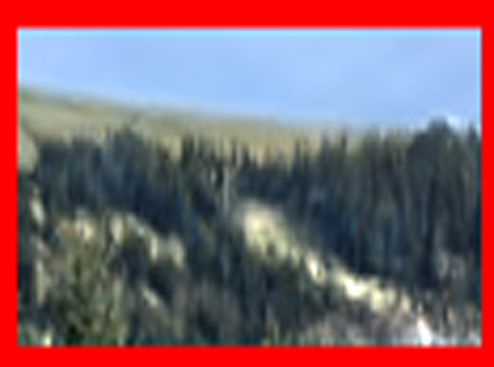}
\includegraphics[width=0.0775\textwidth]{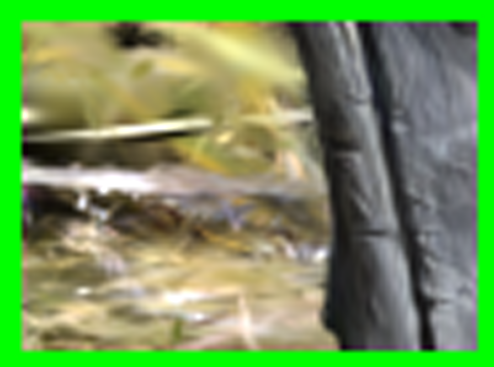}
\includegraphics[width=0.0775\textwidth]{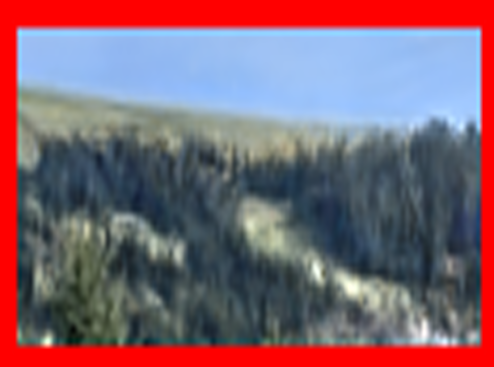}
\includegraphics[width=0.0775\textwidth]{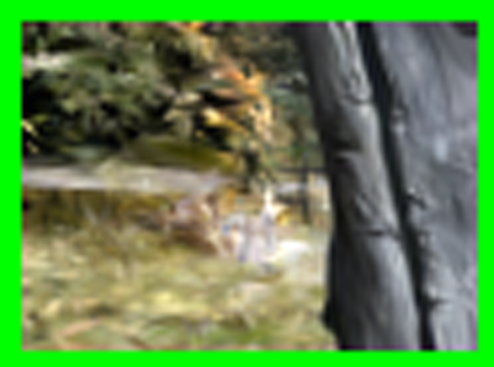}
\\
\includegraphics[width=0.16\textwidth]{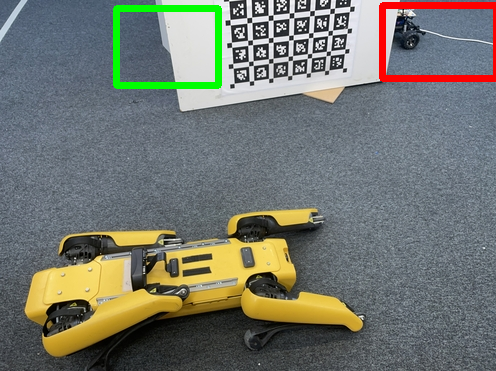}
\includegraphics[width=0.16\textwidth]{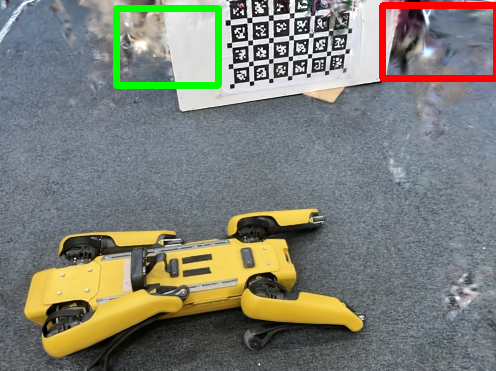}
\includegraphics[width=0.16\textwidth]{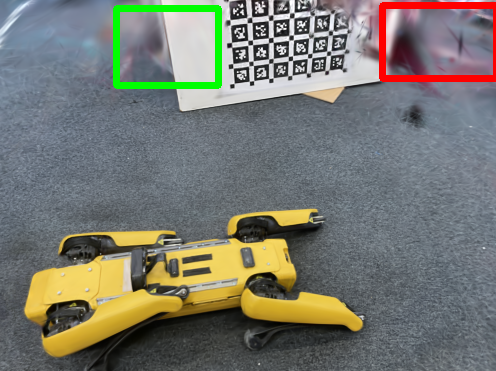}
\includegraphics[width=0.16\textwidth]{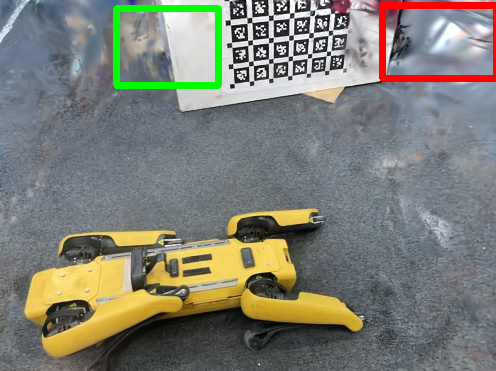}
\includegraphics[width=0.16\textwidth]{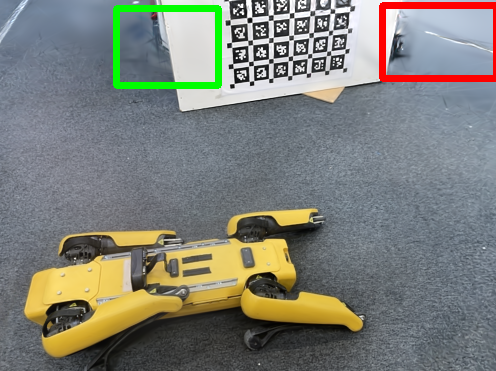}
\includegraphics[width=0.16\textwidth]{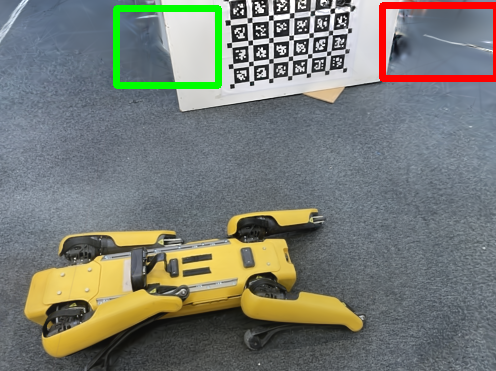}
\\
\color{white}{.}\includegraphics[width=0.0775\textwidth]{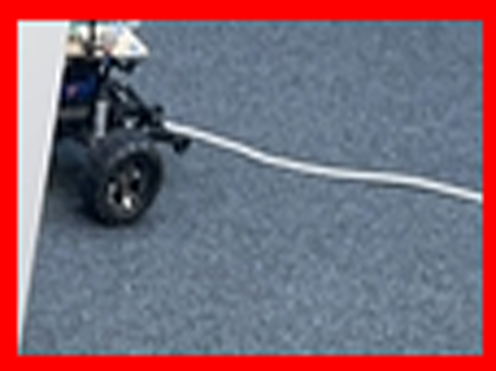}
\includegraphics[width=0.0775\textwidth]{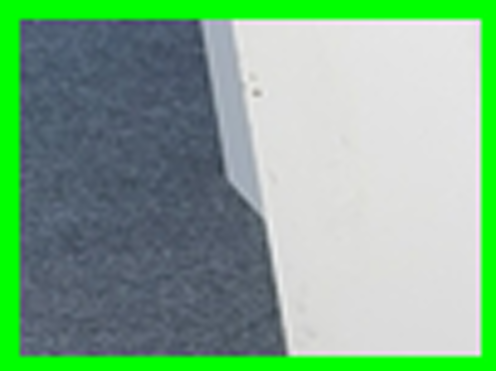}
\includegraphics[width=0.0775\textwidth]{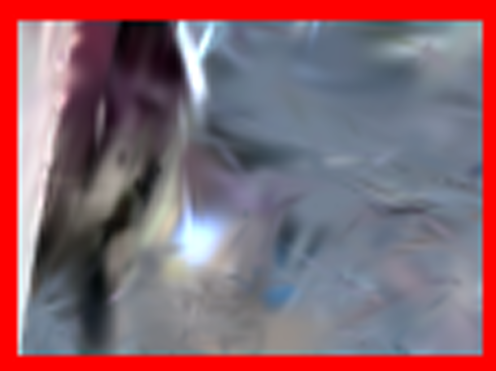}
\includegraphics[width=0.0775\textwidth]{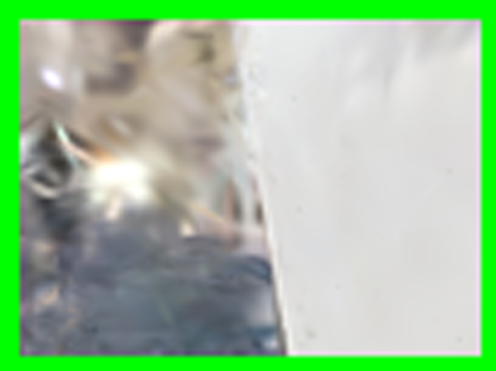}
\includegraphics[width=0.0775\textwidth]{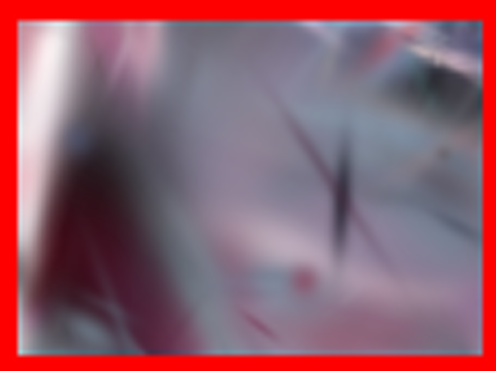}
\includegraphics[width=0.0775\textwidth]{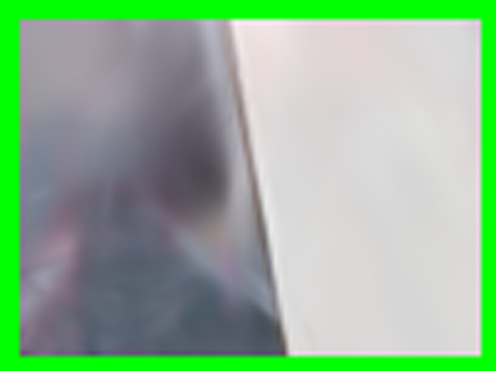}
\includegraphics[width=0.0775\textwidth]{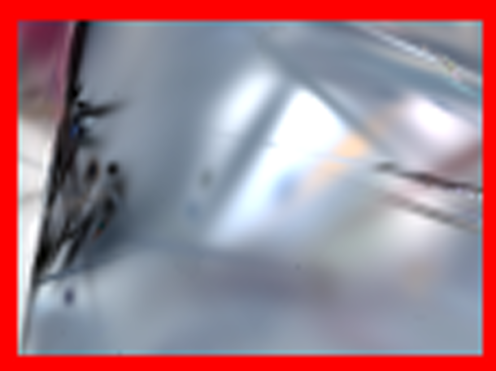}
\includegraphics[width=0.0775\textwidth]{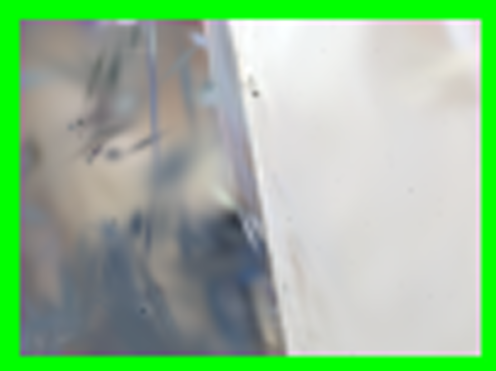}
\includegraphics[width=0.0775\textwidth]{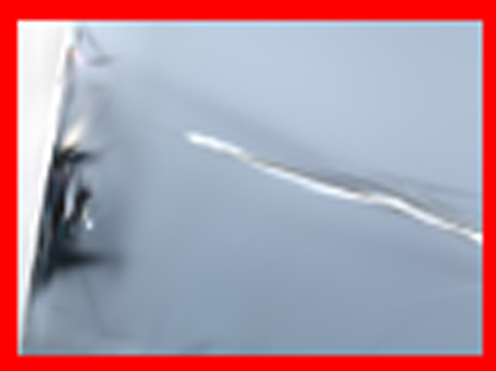}
\includegraphics[width=0.0775\textwidth]{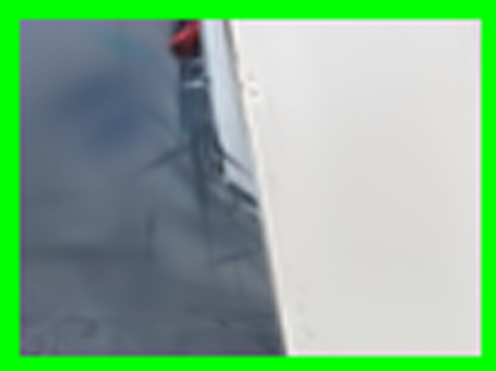}
\includegraphics[width=0.0775\textwidth]{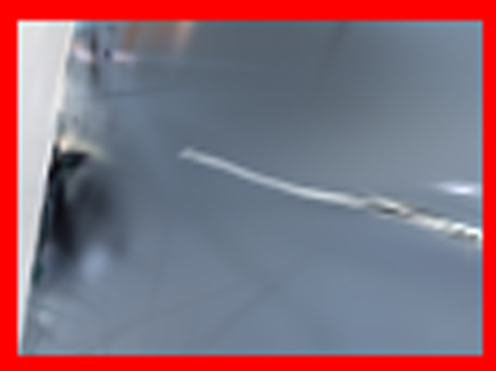}
\includegraphics[width=0.0775\textwidth]{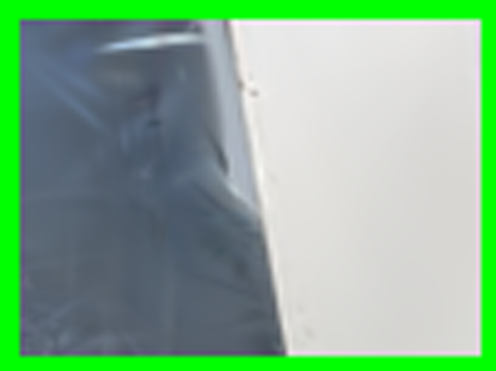}

    \captionof{figure}{Qualitative results on Spot and Mountain from the NeRF On-the-go dataset.}
    \label{fig:on-the-go}
    \vspace{-1em}
\end{table*}

\subsection{Stage II: Prior-Guided Reconstruction}
\label{sec:stage2}

Stage~II reconstructs the scene under the pseudo-mask priors from Sec.~\ref{sec:prior}. 
Its role is not to rediscover transients from scratch, but to \emph{refine} the high-recall pseudo-masks once geometry becomes more stable. 

\paragraph{Prior-guided online mask refinement.}
The pseudo-masks from Sec.~\ref{sec:prior} already provide strong transient suppression, but they may over-mask static regions that are sparsely observed or hard to fit. 
We therefore introduce a lightweight per-pixel MLP that predicts a transient probability map online during the second reconstruction:
\begin{equation}
M_i = \mathrm{MLP}_{\mathrm{mask}}(f_i, d_i),
\end{equation}
where $f_i$ denotes cached image features from the ground-truth training view, and $d_i$ denotes the depth residual between the monocular depth prediction with Depth Anything v2 \cite{yang2024depth} and the current rendered depth. 
Depth residuals provide complementary geometric cues, since transient objects often violate multi-view depth consistency even when their appearance is locally plausible.

\paragraph{Prior supervision before self-consistency.}
We bias the MLP toward the pseudo-masks early in training:
\begin{equation}
\mathcal{L}_{\mathrm{prior}}
=
\exp\!\left(-\frac{t}{\beta_{\mathrm{prior}}}\right)
\left\| M_{\mathrm{pseudo}} - M_i \right\|_1 ,
\end{equation}
where $\beta_{\mathrm{prior}}$ controls the decay rate. 
This stage is intentionally conservative: false positives in $M_{\mathrm{pseudo}}$ only reduce local supervision, whereas false negatives would let transients leak into geometry. 
Strong early prior guidance therefore helps prevent the overfitting-induced signal loss that affects purely online methods. 

\paragraph{Self-consistency after geometry stabilizes.}
As reconstruction stabilizes, supervision gradually shifts from priors to self-consistency cues. 
Following RobustSplat \cite{fu2025robustsplat} refinement, we constrain $M_i$ using residual bounds and feature consistency:
\begin{align}
\mathcal{L}_{\text{res}} &= \max\!\big(U - M_i,\,0\big) + \max\!\big(M_i - L,\,0\big), \label{eq:Lres}\\
\mathbf{M}_{\cos} &= \max\!\big(2\,\cos(f_i, f'_i) - 1,\, 0\big), \label{eq:Mcos}\\
\mathcal{L}_{\cos} &= \big\lVert \mathbf{M}_{\cos} - M_i \big\rVert_1. \label{eq:Lcos}
\end{align}
Concretely, $f_i$ is the cached feature of the ground-truth training view, and $f_i'$ is computed from the current rendering during optimization. We use DINOv2 \cite{oquab2023dinov2} as the feature extraction backbone.
These loss functions are combined as:
\begin{equation}
\mathcal{L}_{\mathrm{robust}}
=
\exp\!\left(
-\frac{\max(0,\,T_{\mathrm{densify}}-t)}{\beta_{\mathrm{robustness}}}
\right)
\left(\mathcal{L}_{\mathrm{cos}}+\mathcal{L}_{\mathrm{res}}\right).
\end{equation}
The final MLP objective is
\begin{equation}
\mathcal{L}_{\mathrm{MLP}}
=
\lambda_{\mathrm{robust}} \mathcal{L}_{\mathrm{robust}}
+
\lambda_{\mathrm{prior}} \mathcal{L}_{\mathrm{prior}}
+
\mathcal{L}_{\mathrm{reg}}.
\end{equation}

In this way, the pseudo-masks act as a strong but temporary prior, while later iterations refine them into an adaptive estimator. 

\section{Experiments}
\label{sec:exp}


\begin{table*}[!th] \centering
    \begin{center}
    \captionof{table}{\textbf{Comparison on RobustNeRF Dataset.} For better visualization, the \colorbox{myred}{1st}, \colorbox{myorange}{2nd} and \colorbox{myyellow}{3rd} best results are highlighted. }

\resizebox{\textwidth}{!}{
\begin{tabular}{l|*{3}{c}|*{3}{c}|*{3}{c}|*{3}{c}|*{3}{c}|*{3}{c}}
    \toprule
    & \multicolumn{3}{c|}{Android} 
    & \multicolumn{3}{c|}{Crab1}
    & \multicolumn{3}{c|}{Crab2}
    & \multicolumn{3}{c|}{Statue}
    & \multicolumn{3}{c|}{Yoda}
    & \multicolumn{3}{c}{Mean}
    \\
    Method
    & \multicolumn{1}{c}{PSNR} 
    & \multicolumn{1}{c}{SSIM} 
    & \multicolumn{1}{c|}{LPIPS} 
    & \multicolumn{1}{c}{PSNR} 
    & \multicolumn{1}{c}{SSIM} 
    & \multicolumn{1}{c|}{LPIPS} 
    & \multicolumn{1}{c}{PSNR} 
    & \multicolumn{1}{c}{SSIM} 
    & \multicolumn{1}{c|}{LPIPS} 
    & \multicolumn{1}{c}{PSNR} 
    & \multicolumn{1}{c}{SSIM} 
    & \multicolumn{1}{c|}{LPIPS} 
    & \multicolumn{1}{c}{PSNR} 
    & \multicolumn{1}{c}{SSIM} 
    & \multicolumn{1}{c|}{LPIPS} 
    & \multicolumn{1}{c}{PSNR} 
    & \multicolumn{1}{c}{SSIM} 
    & \multicolumn{1}{c}{LPIPS} 
    \\
    \midrule
    3DGS~\cite{kerbl20233d}
    & 23.38 & 0.799 & 0.071
    & 31.23 & 0.944 & 0.035
    & 30.34 & 0.925 & 0.055
    & 21.35 & 0.838 & 0.102
    & 30.83 & 0.924 & 0.058
    & 27.43 & 0.886 & 0.064\\
    
    WildGaussians~\cite{kulhanek2024wildgaussians}
    & 23.68 & 0.805 & 0.085
    & 27.20 & 0.916 & 0.065
    & 28.99 & 0.912 & 0.066
    & 22.59 & \cellcolor{myred}0.864 & 0.089
    & 27.19 & 0.915 & 0.071
    & 25.93 & 0.882 & 0.075\\
    
    DeSplat~\cite{wang2025desplat}
    & 24.31 & 0.814 & 0.098
    & 35.63 & \cellcolor{myred}0.969 & 0.045
    & \cellcolor{myyellow}34.91 & 0.950 & 0.099
    & \cellcolor{myorange}22.83 & 0.848 & 0.105
    & 34.34 & 0.953 & 0.095
    & \cellcolor{myyellow}30.41 & \cellcolor{myyellow}0.907 & 0.089\\
    
    SpotLessSplats~\cite{sabour2025spotlesssplats}
    & 24.40 & 0.805 & 0.067
    & 33.29 & 0.940 & 0.034
    & 34.37 & 0.947 & 0.038
    & 22.25 & 0.819 & 0.092
    & \cellcolor{myyellow}34.66 & 0.954 & 0.035
    & 29.79 & 0.893 & 0.053\\

    DeGauss~\cite{wang2025degauss}
    & \cellcolor{myyellow}24.52 & 0.814 & 0.056
    & \cellcolor{myyellow}35.39 & 0.949 & \cellcolor{myyellow}0.027
    & 34.22 & 0.934 & 0.033
    & 22.05 & 0.825 & 0.076
    & 33.78 & 0.936 & 0.032
    & 29.99 & 0.892 & 0.045\\
    
    RobustSplat~\cite{fu2025robustsplat}
    & \cellcolor{myred}24.74 & \cellcolor{myred}0.826 & \cellcolor{myred}0.054
    & \cellcolor{myred}35.95 & \cellcolor{myorange}0.951 & \cellcolor{myorange}0.025
    & \cellcolor{myorange}35.07 & \cellcolor{myorange}0.954 & \cellcolor{myorange}0.032
    & 22.54 & \cellcolor{myyellow}0.855 & 0.074
    & \cellcolor{myred}35.54 & \cellcolor{myred}0.960 & \cellcolor{myred}0.029
    & \cellcolor{myorange}30.77 & \cellcolor{myorange}0.909 & \cellcolor{myorange}0.043\\
    
    3DGS with pseudo-masks
    & 24.36 & \cellcolor{myyellow}0.815 & \cellcolor{myyellow}0.055
    & 34.03 & 0.950 & 0.027
    & 34.41 & \cellcolor{myyellow}0.953 & \cellcolor{myyellow}0.032
    & \cellcolor{myyellow}22.79 & 0.854 & \cellcolor{myyellow}0.072
    & 34.53 & \cellcolor{myyellow}0.957 & \cellcolor{myyellow}0.032
    & 30.02 & 0.906 & \cellcolor{myyellow}0.044\\
    
    Ours
    & \cellcolor{myorange}24.68 & \cellcolor{myorange}0.826 & \cellcolor{myorange}0.055
    & \cellcolor{myorange}35.86 & \cellcolor{myyellow}0.951 & \cellcolor{myred}0.024
    & \cellcolor{myred}35.08 & \cellcolor{myred}0.955 & \cellcolor{myred}0.031
    & \cellcolor{myred}23.09 & \cellcolor{myorange}0.863 & \cellcolor{myred}0.069
    & \cellcolor{myorange}35.42 & \cellcolor{myorange}0.960 & \cellcolor{myorange}0.030
    & \cellcolor{myred}30.83 & \cellcolor{myred}0.911 & \cellcolor{myred}0.042\\
    \bottomrule
\end{tabular}
}

    \label{table:robustnerf}
    \end{center}
    \vspace{-15pt}
    \makebox[0.16\textwidth]{\footnotesize GT}
\makebox[0.16\textwidth]{\footnotesize DeSplat}
\makebox[0.16\textwidth]{\footnotesize WildGaussians}
\makebox[0.16\textwidth]{\footnotesize SpotLessSplats}
\makebox[0.16\textwidth]{\footnotesize RobustSplat}
\makebox[0.16\textwidth]{\footnotesize Ours}
\\
\includegraphics[width=0.16\textwidth]{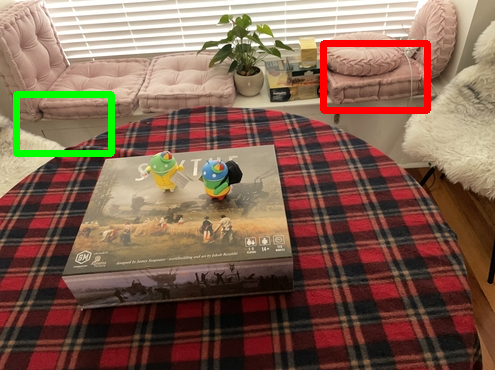}
\includegraphics[width=0.16\textwidth]{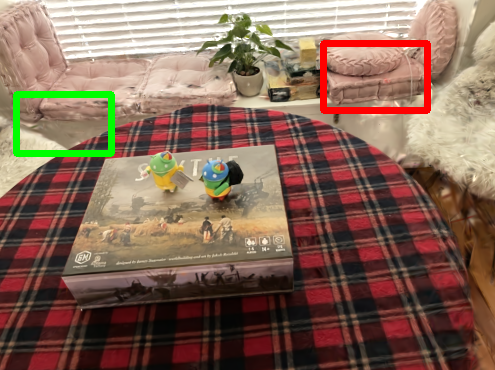}
\includegraphics[width=0.16\textwidth]{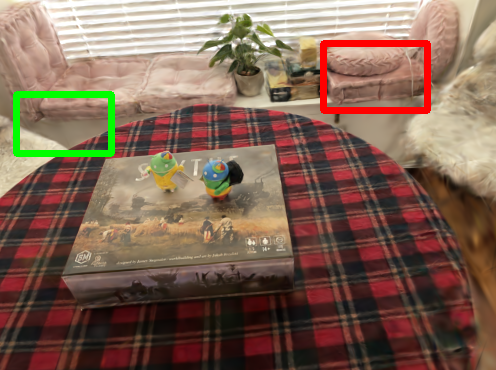}
\includegraphics[width=0.16\textwidth]{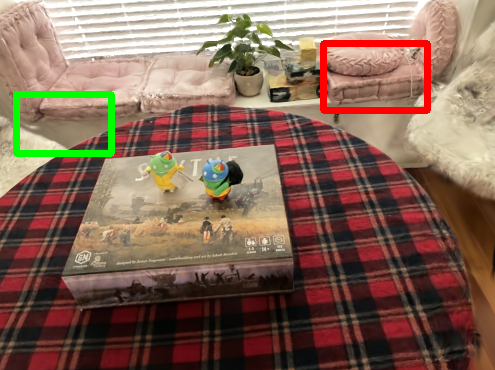}
\includegraphics[width=0.16\textwidth]{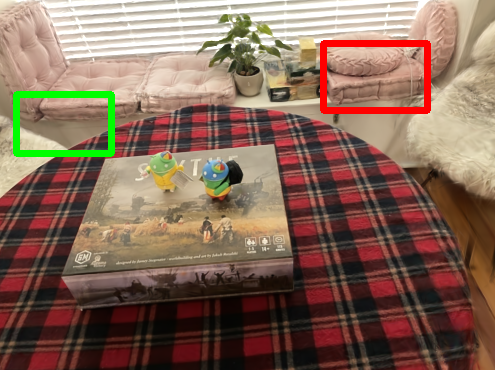}
\includegraphics[width=0.16\textwidth]{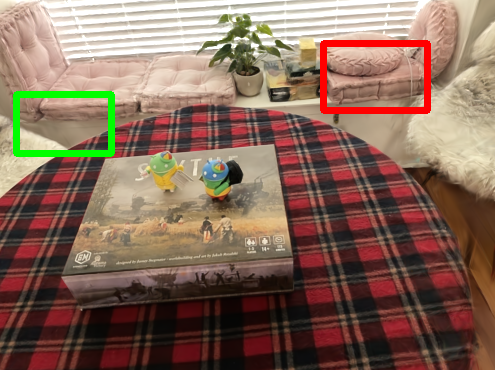}
\\
\includegraphics[width=0.0775\textwidth]{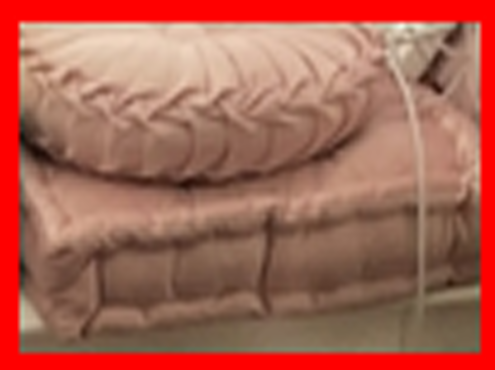}
\includegraphics[width=0.0775\textwidth]{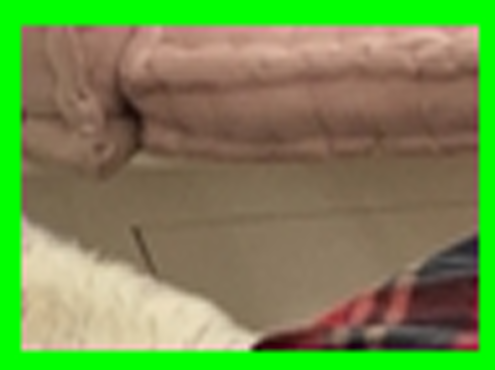}
\includegraphics[width=0.0775\textwidth]{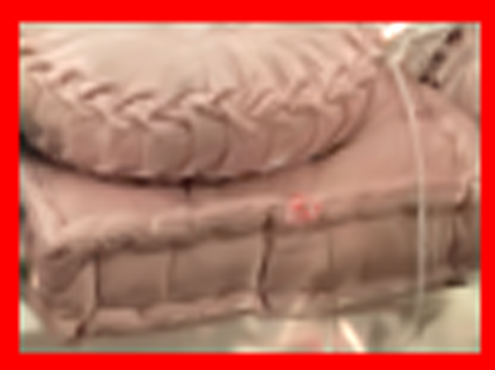}
\includegraphics[width=0.0775\textwidth]{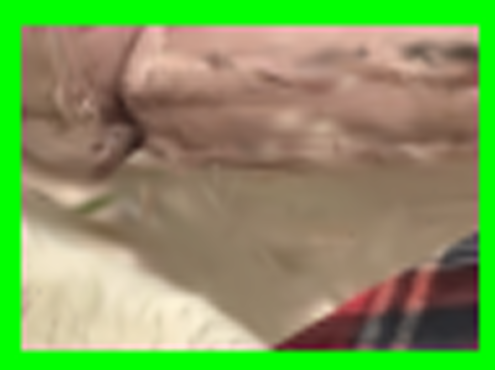}
\includegraphics[width=0.0775\textwidth]{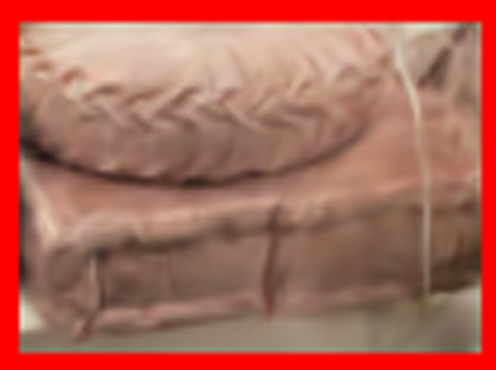}
\includegraphics[width=0.0775\textwidth]{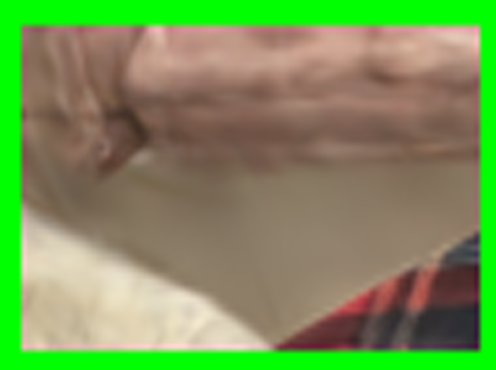}
\includegraphics[width=0.0775\textwidth]{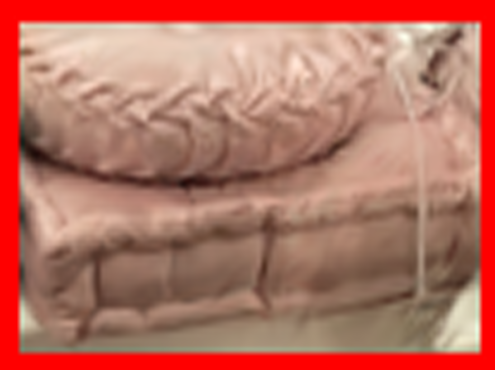}
\includegraphics[width=0.0775\textwidth]{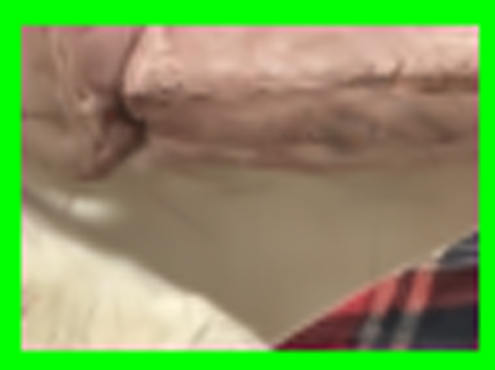}
\includegraphics[width=0.0775\textwidth]{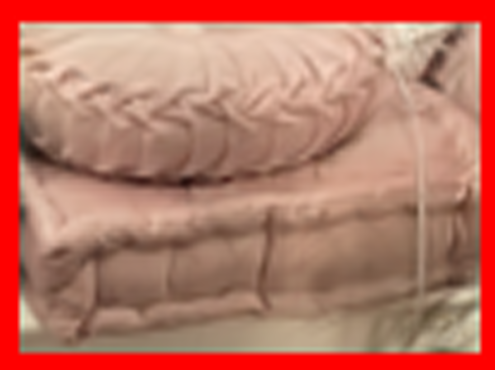}
\includegraphics[width=0.0775\textwidth]{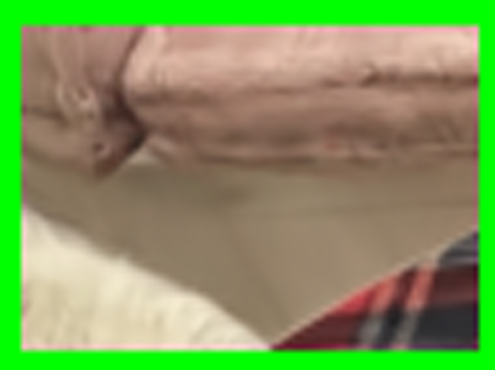}
\includegraphics[width=0.0775\textwidth]{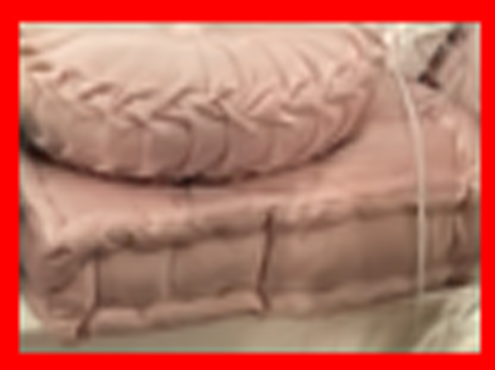}
\includegraphics[width=0.0775\textwidth]{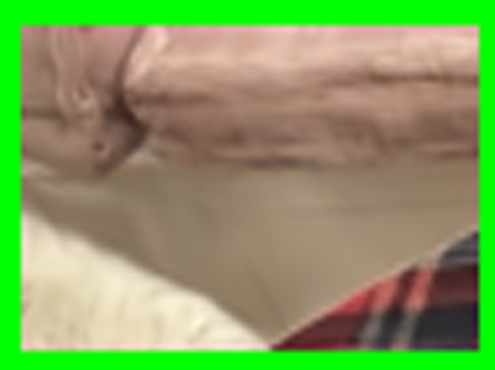}
\\
\includegraphics[width=0.16\textwidth]{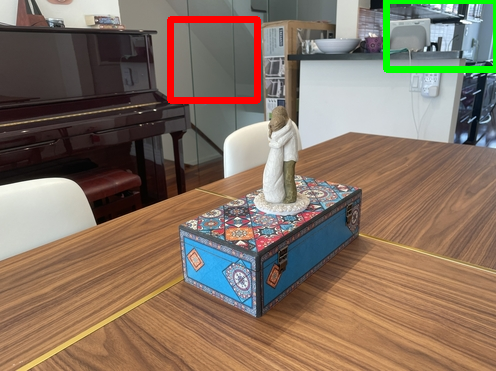}
\includegraphics[width=0.16\textwidth]{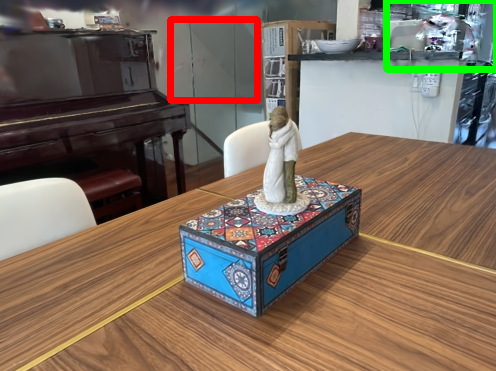}
\includegraphics[width=0.16\textwidth]{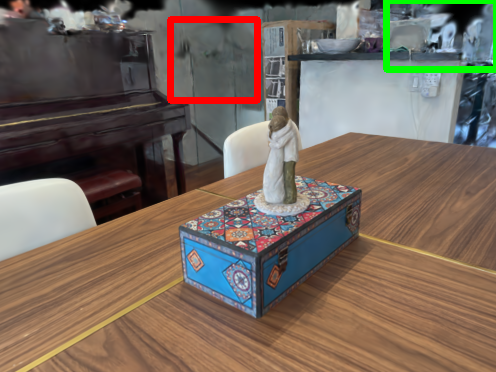}
\includegraphics[width=0.16\textwidth]{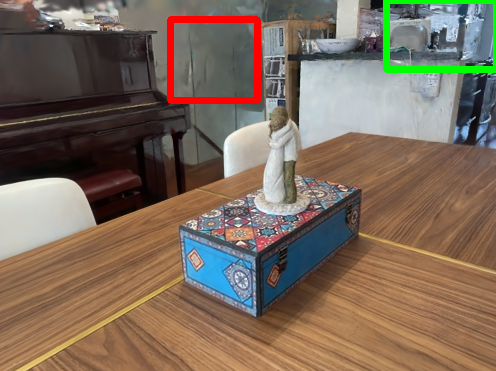}
\includegraphics[width=0.16\textwidth]{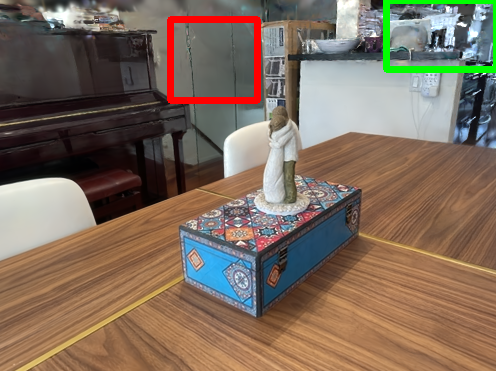}
\includegraphics[width=0.16\textwidth]{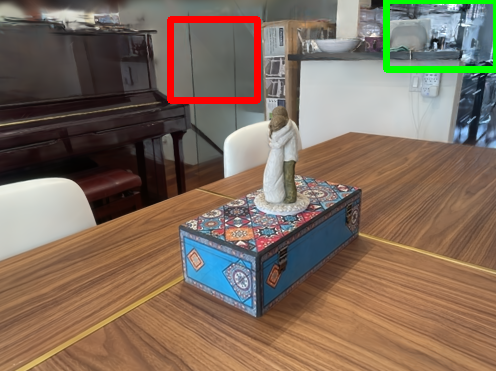}
\\
\color{white}{.}\includegraphics[width=0.0775\textwidth]{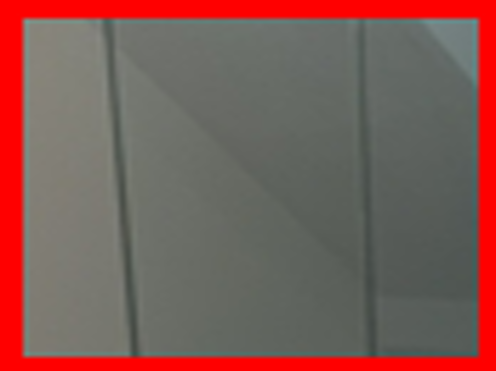}
\includegraphics[width=0.0775\textwidth]{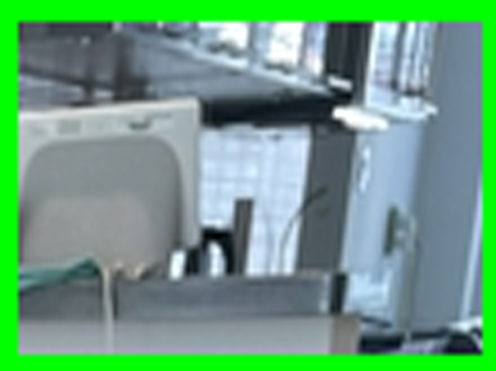}
\includegraphics[width=0.0775\textwidth]{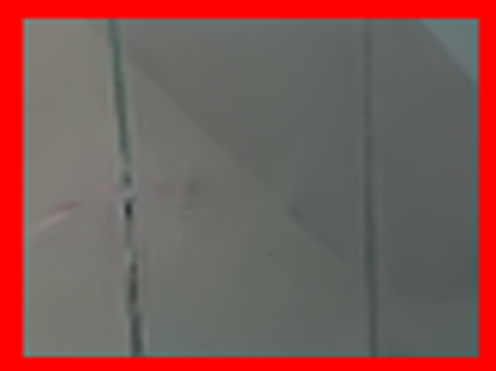}
\includegraphics[width=0.0775\textwidth]{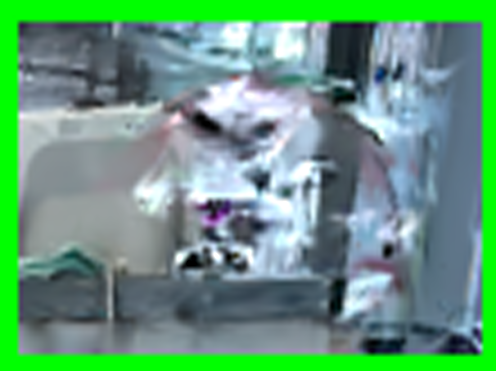}
\includegraphics[width=0.0775\textwidth]{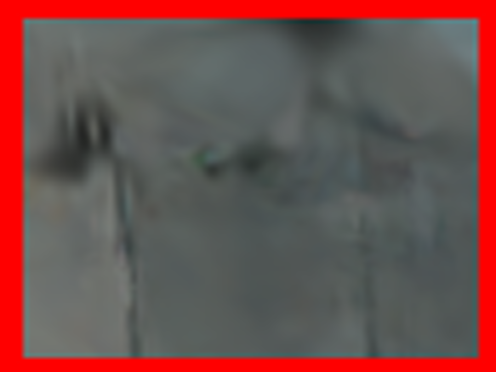}
\includegraphics[width=0.0775\textwidth]{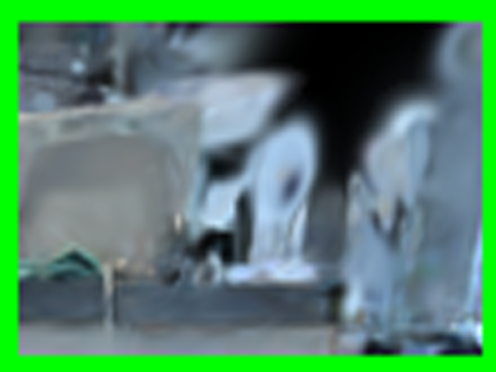}
\includegraphics[width=0.0775\textwidth]{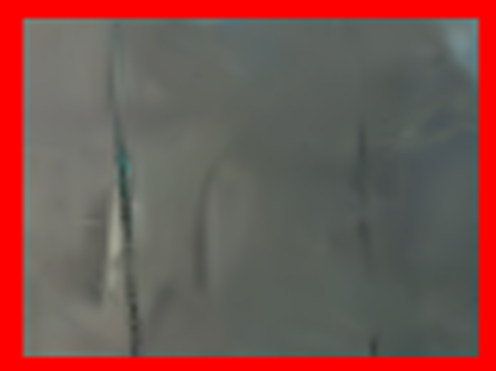}
\includegraphics[width=0.0775\textwidth]{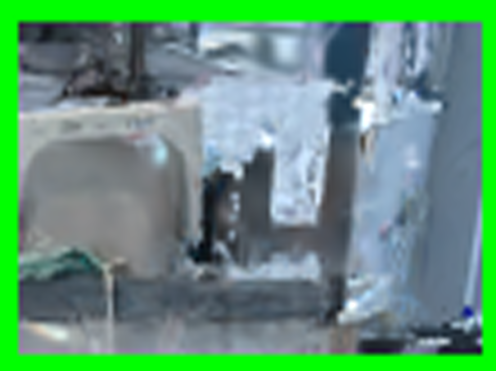}
\includegraphics[width=0.0775\textwidth]{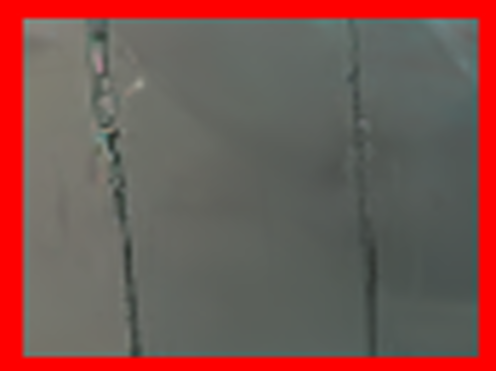}
\includegraphics[width=0.0775\textwidth]{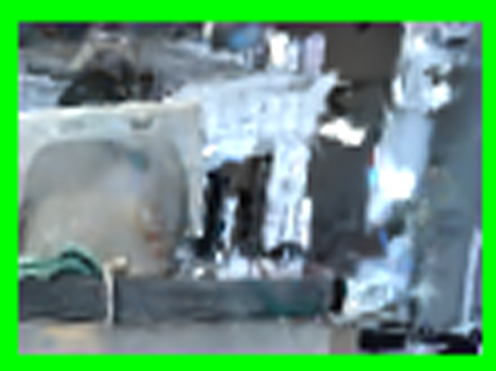}
\includegraphics[width=0.0775\textwidth]{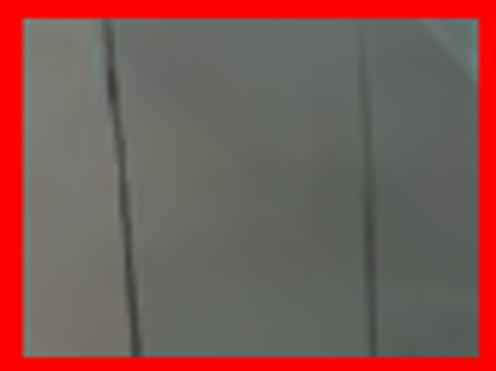}
\includegraphics[width=0.0775\textwidth]{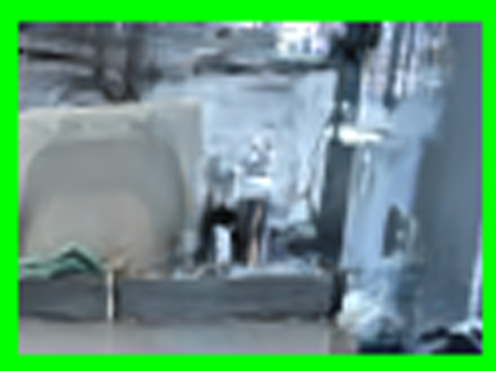}
    \captionof{figure}{Qualitative results on Statue and Android from the RobustNeRF dataset.}
    \label{fig:robustnerf}
    \vspace{-1em}
\end{table*}

\subsection{Setups}
We evaluated \modelName~on two standard datasets for transient-free reconstruction: RobustNeRF  \cite{sabour2023robustnerf} and NeRF On-the-go \cite{ren2024nerf}. These datasets contain diverse outdoor scenes with varying transient densities, enabling a comprehensive assessment of robustness and reconstruction quality. In \cref{sec:compare}, we compare our method against 3DGS-based baselines using both quantitative metrics and qualitative visualizations. \cref{sec:ablation} presents ablation studies to validate the contribution of each core component in handling occlusions and improving overall reconstruction quality.
 
\paragraph{Datasets.}We evaluate \modelName~on the RobustNeRF \cite{sabour2023robustnerf} and NeRF On-the-go \cite{ren2024nerf} datasets. For RobustNeRF, we conduct experiments on all five available scenes (\textit{Statue, Android, Crab (1), Crab (2)}, and \textit{Yoda}), and downsample every image by 8×. For NeRF On-the-go, we evaluate on six scenes: \textit{Mountain, Fountain, Corner, Patio, Spot, and Patio-High}. All images are downsampled by 8× except for Patio, which uses 4×.

\paragraph{Baselines.}We compared \modelName~with recent 3DGS-based transient-suppression methods, including SpotLessSplats \cite{sabour2025spotlesssplats}, WildGaussians \cite{kulhanek2024wildgaussians}, DeSplat \cite{wang2025desplat}, DeGauss \cite{wang2025degauss} and RobustSplat \cite{fu2025robustsplat}. To ensure a fair comparison among the 3DGS approaches, we train all models from the same point clouds. We additionally include a 3DGS \cite{kerbl20233d} variant that directly applies the pseudo-masks without any additional refinement. Evaluation follows standard image metrics—PSNR, SSIM, LPIPS.

\paragraph{Implementation Details.} We implement DualSplat in PyTorch based on RobustSplat \cite{fu2025robustsplat}. The first stage runs for \textbf{10k} iterations with the 3DGS \cite{kerbl20233d} default densification settings to quickly obtain an initial conservative reconstruction. We then extract pseudo-masks using the mask filter. The second stage runs for \textbf{30k} iterations, with densification enabled between 10k and 20k iterations to stabilize early optimization. In addition, we perform the second reconstruction with depth regularization. All parameters, including gaussian primitive parameters, are optimized with Adam using the default learning rates, while the MLP uses a learning rate of \(1\times 10^{-3}\). We set \(\lambda_{\text{local}}=1.5\), \(\lambda_{\text{robust}}=0.5,\lambda_\text{prior}=1\), \(T_{\text{densify}}=10{,}000\), \(\beta_{\text{robustness}}=10{,}000\), and \(\beta_{\text{prior}}=10{,}000\). For mask filter, we set \(\tau_\text{sim} = 0.75\) and \(\tau_{L_1}=0.05\). We inherit RobustSplat’s progressive MLP training schedule and other hyperparameters in Stage II unless otherwise stated.

\subsection{Distractor-free 3D Reconstruction}
\label{sec:compare}

\paragraph{Comparison on NeRF On-the-go Dataset.} 

\cref{table:on-the-go} reports the quantitative comparison on NeRF On-the-go. 
\modelName~achieves the best overall average performance. 
Compared with vanilla 3DGS, the gain is substantial, confirming that transient-aware masking is essential for in-the-wild 3DGS reconstruction. 
These results demonstrate that our method provides a better balance between transient suppression and static-detail preservation across different occlusion levels.

The qualitative comparisons in \cref{fig:on-the-go} further support this trend. 
On the Mountain and Spot scenes, several baselines either suffer from noticeable background blurring or distractor-induced artifacts in the highlighted regions, whereas \modelName~reconstructs cleaner background structures and preserves sharper details around the tree boundary and distant scene content.

\paragraph{Comparison on RobustNeRF Dataset.} 



We further evaluate \modelName~on RobustNeRF, with the quantitative results summarized in \cref{table:robustnerf}. 
\modelName~again achieves the best average performance, slightly surpassing other baselines on all three mean metrics. 
Although the margins are modest and some individual scenes are still led by competing methods, our method remains consistently competitive across all five scenes and achieves the strongest overall balance between reconstruction fidelity and distractor suppression. 

As shown in \cref{fig:robustnerf}, the qualitative differences are clearer than what the averaged metrics alone suggest. 
In the upper example, \modelName~recovers the background texture as well as other baselines. 
In the lower example, our method reduces erroneous background occlusion more effectively and reconstructs a cleaner static layout around the object and wall boundary. 
These visual results indicate that DualSplat not only removes transient artifacts, but also better avoids the over-suppression of static regions.

\subsection{Ablation Study}
\label{sec:ablation}


We conducted comprehensive ablation studies to validate the contribution of each component. Unless otherwise stated, all experiments are performed on the NeRF On-the-go dataset. We report averaged metrics with relative gains over vanilla 3DGS baseline. All ablations are retrained from the same initialization and schedule to ensure fair comparison.


\paragraph{Effects of model modules.}
\begin{table}[h]
  \centering
  \renewcommand{\arraystretch}{1.05} 
  \caption{Comparison of different modules.}
  \vspace{-1em}
  \resizebox{\linewidth}{!}{
    \begin{tabular}{l|ccc}
    \toprule
    \multirow{1}{*}{Method} 
    & \multicolumn{3}{c}{Mean}  
    \\
     & PSNR & SSIM & LPIPS 
    \\
    \midrule
    base (3DGS)             & 19.043 & 0.697 & 0.196 \\
    base+PM                 & 22.604 & 0.810 & 0.089 \\ \hline
    DD                      & 20.820 & 0.764 & 0.145 \\
    DD+PM                   & 22.899 & 0.818 & 0.090 \\ \hline
    DD+MLP w/o robust loss   & 22.902 & 0.817 & 0.090 \\
    DD+MLP w/o pseudo-masks & 23.122 & 0.818 & 0.089 \\ \hline
    DD+MLP                  & 23.262 & 0.820 & 0.088 \\ 
    DD+MLP + depth regularization             & \textbf{23.421} & \textbf{0.820} & \textbf{0.088}\\
    \bottomrule
    \end{tabular}}
  \label{table:effect-module}
  \vspace{-1em}
  \end{table}


We further decompose \modelName~into three main components and perform controlled ablations: (i) Delayed Densification (DD) for 3DGS; (ii) pseudo-mask application (PM), which directly applies the filtered pseudo-masks without any learned correction; and (iii) the MLP mask predictor, evaluated under two variants—MLP w/o pseudo-masks (trained only with feature/residual cues, no pseudo-mask supervision) and MLP w/o robustness loss (pseudo-mask supervision kept, but the robustness loss removed). We also report the effect of depth regularization on the evaluation metrics.

As summarized in \cref{table:effect-module}, the PM module contributes the largest performance gain (+3.56 dB), demonstrating that accurate transient filtering is the key to the success of our method. The combination of DD and PM further provides an additional improvement (+0.29 dB), confirming that delaying densification allows for more reliable transient suppression before scene geometry is finalized. Moreover, the proposed MLP module plays a crucial role in the refinement of transient masks, generating a notable improvement (+0.37 dB). 
Collectively, these findings validate that each component targets distinct challenges in transient-aware reconstruction. Furthermore, their synergistic integration is crucial for achieving the robustness and high fidelity observed in our final model.

\paragraph{Stratified transient/static analysis.}

To further clarify whether the gain of \modelName~comes from improved transient robustness or generic reconstruction sharpening, we additionally compare \modelName~with RobustSplat\cite{fu2025robustsplat} on transient regions (Inside) and static background (Outside). As shown in \cref{tab:stratified}, the advantage of DualSplat is concentrated in transient regions, while its performance on static background remains comparable to RobustSplat. This indicates that the improvement mainly comes from better artifact removal enabled by our design, rather than from background smoothing or depth regularization alone.

\begin{table}[t]
  \centering
  \renewcommand{\arraystretch}{1} 
  \caption{Stratified performance: Transient vs. Static regions.}\vspace{-1em}
  \resizebox{\linewidth}{!}{
      \begin{tabular}{l|ccc|ccc|ccc}
        \toprule
        Transient Regions 
        & \multicolumn{3}{c|}{Crab1}  & \multicolumn{3}{c|}{Crab2} & \multicolumn{3}{c}{Yoda}
        \\
         (Inside)& PSNR & SSIM & LPIPS & PSNR & SSIM & LPIPS & PSNR & SSIM & LPIPS 
        \\
        \midrule
        RobustSplat             & 32.78          & 0.929          & \textbf{0.0065} & \textbf{35.58} & 0.954 & 0.0025          & 35.45          & 0.961          & \textbf{0.0028} \\
        Ours                    & \textbf{33.04} & \textbf{0.931} & 0.0074          & 35.55          & \textbf{0.956} & \textbf{0.0022} & \textbf{35.57} & \textbf{0.962} & 0.0029          \\ 
        \bottomrule
        \end{tabular}
    }
    \resizebox{\linewidth}{!}{
      \begin{tabular}{l|ccc|ccc|ccc}
        \toprule
        Static Background
        & \multicolumn{3}{c|}{Crab1}  & \multicolumn{3}{c|}{Crab2} & \multicolumn{3}{c}{Yoda}
        \\
        (Outside) & PSNR & SSIM & LPIPS & PSNR & SSIM & LPIPS & PSNR & SSIM & LPIPS 
        \\
        \midrule
        RobustSplat             & \textbf{36.38} & \textbf{0.953} & 0.0212          & 35.71          & 0.958          & 0.0257          & 36.01          & \textbf{0.963} & \textbf{0.0236} \\
        Ours                    & 36.36          & \textbf{0.953} & \textbf{0.0210} & \textbf{35.75} & \textbf{0.959} & \textbf{0.0253} & \textbf{36.05} & \textbf{0.963} & 0.0241          \\ 
        \bottomrule
        \end{tabular}
    }
    \label{tab:stratified}\vspace{-1em}
\end{table}

\paragraph{Effects of feature extractor.} 

\begin{table}[t]
  \centering
  \caption{Comparison of different feature extraction models.}
  \vspace{-1em}
  \resizebox{\linewidth}{!}{
    \begin{tabular}{l|cccc}
    \toprule
    Methods & Accuracy & Precision & Recall & IoU
    \\
    \midrule
    Ours*	& \textbf{0.988}&	\textbf{0.863}&	0.950&	\textbf{0.858} \\
    FiT3D \cite{yue2024improving}      & 0.976    & 0.841     & 0.951  & 0.835 \\
    DINOv2 \cite{oquab2023dinov2}       & 0.945    & 0.747     & \textbf{0.956}  & 0.744 \\
    SD \cite{rombach2022high}         & 0.942    & 0.741     & 0.951  & 0.738 \\
    ResNet18 \cite{he2016deep}      & 0.923    & 0.655     & 0.708  & 0.561 \\
    ResNet50 \cite{he2016deep}      & 0.925    & 0.679     & 0.798  & 0.616 \\
    ResNet101 \cite{he2016deep}     & 0.929    & 0.682     & 0.788  & 0.615 \\
    VGG16 \cite{simonyan2014very}      & 0.907    & 0.628     & 0.706  & 0.540 \\
    VGG19 \cite{simonyan2014very}      & 0.912    & 0.631     & 0.683  & 0.533 \\ 
    \bottomrule
    \end{tabular}}
    \vspace{-1em}
  \label{table:effect-mask}

\end{table}

To assess the impact of the feature extractor on the mask filter, we compare several backbones under the same pipeline. The ground-truth transient masks on six NeRF On-the-go scenes are bootstrapped from SAM2 \cite{ravi2024sam} proposals and manually curated. We then evaluated DINOv2 \cite{oquab2023dinov2}, Stable Diffusion \cite{rombach2022high}, FiT3D \cite{yue2024improving}, and classic CNNs (ResNet \cite{he2016deep}, VGG \cite{simonyan2014very}) with shared preprocessing, similarity computation, thresholds, and post-processing. We apply the same evaluation protocol to the masks predicted by our MLP. We treat the predicted mask as a binary feature-consistency map, and apply the mask-filtering step to derive the final mask, denoting this variant as Ours*. As reported in \cref{table:effect-mask}, FiT3D yields better transient masks than other pretrained models, especially in precision and IoU, which are key to accurately delineating object boundaries and preventing transient leakage during training. Our method further improves pseudo-masks in accuracy, precision and recall.
\section{Conclusion}

We introduced \modelName, a robust framework for transient-free view synthesis. \modelName~ builds reliable pseudo-masks from mismatches between captured images and intermediate renderings, and uses them to mask dynamic content from the static reconstruction. Comprehensive experiments on challenging benchmarks show improvements in robustness over recent 3DGS-based baselines, with ablations attributing the improvements to each component of the pipeline.

\paragraph{Limitations.}
\modelName~increases overall training time due to its two-stage design and SAM2 mask generation, and the MLP is trained per scene without explicit generalization. We also observe that the MLP predictor produces false positives, especially near object boundaries. However, since transient masking is asymmetric, these errors have a negligible impact on densely observed static regions. In our tests on private data, we also observe a failure mode: when a transient object persists for a long duration and is visible in a substantial fraction of the views, it becomes harder to suppress.

\section*{Acknowledgments}
This work was supported by the National Natural Science Foundation of China under Grant 62388101, the National Key R\&D Program of China (No. 2023YFB3309000), and the Fundamental Research Funds for Central Universities under Grant YWF-22-L-1281.

{
    \small
    \bibliographystyle{ieeenat_fullname}
    \bibliography{main}
}


\end{document}